\documentclass[runningheads]{llncs}  

\usepackage{graphicx}

\usepackage{tikz}
\usepackage{comment}
\usepackage{amsmath,amssymb} 
\usepackage{color}

\usepackage[accsupp]{axessibility}  

\usepackage{hhline}
\usepackage{multirow}
\usepackage{makecell}
\usepackage{graphicx}
\usepackage{paralist}
\usepackage{appendix}
\usepackage{algorithm}
\usepackage{float}
\usepackage{algpseudocode}

\usepackage{wrapfig}

\makeatletter
\newcommand\footnoteref[1]{\protected@xdef\@thefnmark{\ref{#1}}\@footnotemark}
\makeatother

\usepackage{xcolor,colortbl}
\newcommand{\first}[1]{\cellcolor{yellow!75}\textbf{#1}}
\newcommand{\second}[1]{\cellcolor{yellow!25}{#1}}

\definecolor{gray}{rgb}{0.5, 0.5, 0.5}
\definecolor{darkgray}{rgb}{0.2, 0.2, 0.2}
\definecolor{scarlet}{rgb}{1, 0.35, 0.15}
\definecolor{mulberry}{rgb}{0.773, 0.294, 0.549}
\definecolor{blue}{rgb}{0, 0, 1}
\definecolor{skyblue}{rgb}{0.3, 0.7, 0.9}
\definecolor{darkgreen}{rgb}{0.2, 0.7, 0.1}
\definecolor{darkyellow}{rgb}{1, 0.65, 0}

\usepackage{xspace}



\newcommand{\cmark}{\ding{51}}%

\usepackage{xcolor,pifont}
\newcommand*\colourcheck[1]{%
  \expandafter\newcommand\csname #1check\endcsname{\textcolor{#1}{\ding{52}}}%
}
\colourcheck{green}

\newcommand*\colourxmark[1]{%
  \expandafter\newcommand\csname #1xmark\endcsname{\textcolor{#1}{\ding{55}}}%
}
\colourxmark{red}

\makeatletter
\DeclareRobustCommand\onedot{\futurelet\@let@token\@onedot}
\def\@onedot{\ifx\@let@token.\else.\null\fi\xspace}
 
\def\ie{\emph{i.e}\onedot}

\def\etal{\emph{et al}\onedot}

\newcommand{\Tref}[1]{Table~\textcolor{blue}{\ref{#1}}}
\newcommand{\Eref}[1]{Eq.~\textcolor{blue}{\ref{#1}}}
\newcommand{\Fref}[1]{Fig.~\textcolor{blue}{\ref{#1}}}
\newcommand{\Sref}[1]{Sec.~\textcolor{blue}{\ref{#1}}}
\newcommand{\Aref}[1]{Alg.~\textcolor{blue}{\ref{#1}}}

\newcommand\blfootnote[1]{%
  \begingroup
  \renewcommand\thefootnote{}\footnote{#1}%
  \addtocounter{footnote}{-1}%
  \endgroup
}

\usepackage[width=122mm,left=12mm,paperwidth=146mm,height=193mm,top=12mm,paperheight=217mm]{geometry}

\begin{document}
\pagestyle{headings}
\mainmatter
\def\ECCVSubNumber{4619}  

\title{Facial Depth and Normal Estimation \\ using Single Dual-Pixel Camera} 

%
%
\authorrunning{Kang et al.}
\author{
  Minjun Kang\textsuperscript{1}$^{\dagger}$ \quad Jaesung Choe\textsuperscript{1} \quad Hyowon Ha\textsuperscript{4} \quad Hae-Gon Jeon\textsuperscript{2} \quad Sunghoon Im\textsuperscript{3} \quad In So Kweon\textsuperscript{1} \quad Kuk-Jin Yoon\textsuperscript{1}\\ \vspace{1mm}
  \textsuperscript{1}KAIST \quad \textsuperscript{2}GIST \quad \textsuperscript{3}DGIST \quad \textsuperscript{4}Meta Reality Labs \\
  \textsuperscript{1}\url{https://github.com/MinJunKang/DualPixelFace}
}

\institute{}

\maketitle
\vspace{-4mm}
\begin{abstract}
Recently, Dual-Pixel (DP) sensors have been adopted in many imaging devices.
However, despite their various advantages, DP sensors are used just for faster auto-focus and aesthetic image captures, and research on their usage for 3D facial understanding has been limited due to the lack of datasets and algorithmic designs that exploit parallax in DP images.
It is also because the baseline of sub-aperture images is extremely narrow, and parallax exists in the defocus blur region. In this paper, we introduce a DP-oriented Depth/Normal estimation network that reconstructs the 3D facial geometry. In addition, to train the network, we collect DP facial data with more than 135K images for 101 persons captured with our multi-camera structured light systems. 
It contains ground-truth 3D facial models including depth map and surface normal in metric scale. Our dataset allows the proposed network to be generalized for 3D facial depth/normal estimation. The proposed network consists of two novel modules: Adaptive Sampling Module (ASM) and Adaptive Normal Module (ANM), which are specialized in handling the defocus blur in DP images. Finally, we demonstrate that the proposed method achieves state-of-the-art performances over recent DP-based depth/normal estimation methods.
\vspace{-1mm}

\end{abstract}

\blfootnote{$\dagger$ Email address : kmmj2005@kaist.ac.kr}

\vspace{-8mm}
\section{Introduction}
\label{sec:intro}
\vspace{-2mm}
A huge number of facial images are posted every day on social media. In 2020, for example, about 70 percent of photos were taken using cameras on smartphones and 24 billion selfies were uploaded to Google Photos App~\cite{Google_stat,Google_stat2}. 
Accordingly, acquiring facial geometry from images 
has emerged as an interesting research topic, since 3D facial geometry can be used for various applications~\cite{zollhofer2018state} such as face recognition \cite{hu2017finding,liu2018joint,liang20203d}, 
performance-based animation, real-time facial reenactment~\cite{Augmented_Faces,zhou2019deep}, facial biometrics~\cite{yu2021deep}, face-based interfaces, visual speech recognition, face-based search in visual assets, creating personalized avatars~\cite{luo2021normalized,lattas2020avatarme} or 3D printing of faces for entertainment or medicine, facial puppetry, face replacement \cite{zhou2019deep,Augmented_Faces}, speech-driven animation, virtual make-up, and face image editing~\cite{yang2011expression}, etc.
%
%
3D facial geometry 
can be obtained by either using multiple cameras~\cite{s10,iPhone11} or active sensing devices~\cite{kinect2,keselman2017intel}. However, these methods often suffer from uncontrolled lighting conditions or hardware synchronization.

%
Recently, \textbf{Dual-Pixel (DP)} sensors get noticed due to their various advantages 
and are applied to the many portable imaging devices 
such as iPhone13 ProMax and Samsung Galaxy 22. DP sensors are perfectly synchronized with the same exposure, white balance, and geometric rectification. Such strengths derive from their hardware configuration that captures two images in a single camera at once. 
Based on these properties, currently, these novel sensors are mainly specialized in the fast auto-focus operation and more aesthetic image captures. 
However, thanks to their characteristics, DP sensors have great potential for other tasks. For example, a few previous studies~\cite{wadhwa2018synthetic,garg2019learning,punnappurath2020modeling,du2net,DP-explore,Xin_2021_ICCV_dual_pixel} envision the new possibility of DP images for scene depth estimation. Usually, these studies regard DP images as extremely narrow-baseline stereo images having defocus-blur to infer high-quality depth maps. 
Here, it is worthy to note that although the DP sensors are being actively used to take face pictures, 
there has been a limited study~\cite{wu2020single} that recovers facial geometry using a Dual-Pixel camera. We found that previous methods have difficulty in facial geometry estimation, which is due to the lack of a facial DP dataset with precise 3D geometry and an appropriate algorithm for generalized estimation.

To address the issue, we present a DP-oriented 3D facial dataset and a depth/normal estimation network toward high-quality facial geometry reconstruction with DP cameras. We represent the 3D facial geometry not only with the depth map but also with the normal map for various applications such as face relighting. 
Our dataset involves 135,744 face data for 101 persons consisting of DP images and their corresponding depth maps and surface normal maps, which are captured by our structured light camera system. Based on these data, we train our depth/normal estimation network, called stereoDPNet, to infer 3D facial information from DP images. In particular, our stereoDPNet is fully oriented from the properties of dual-pixel images that have an extremely small range of disparity with defocus-blur. Our network design carefully treats these distinctive properties through our Adaptive Sampling Module (ASM) and Adaptive Normal Module (ANM). Finally, the contributions are as follows:
\begin{itemize}
    \item DP-oriented 3D facial dataset with more than 135K DP images and their corresponding high-quality 3D models. 
    \item Novel depth/normal estimation network for 
    facial 3D reconstruction from a DP image with better generalization.
\end{itemize}

\begin{figure}[!t]
\centering
\includegraphics[width=0.98\linewidth]{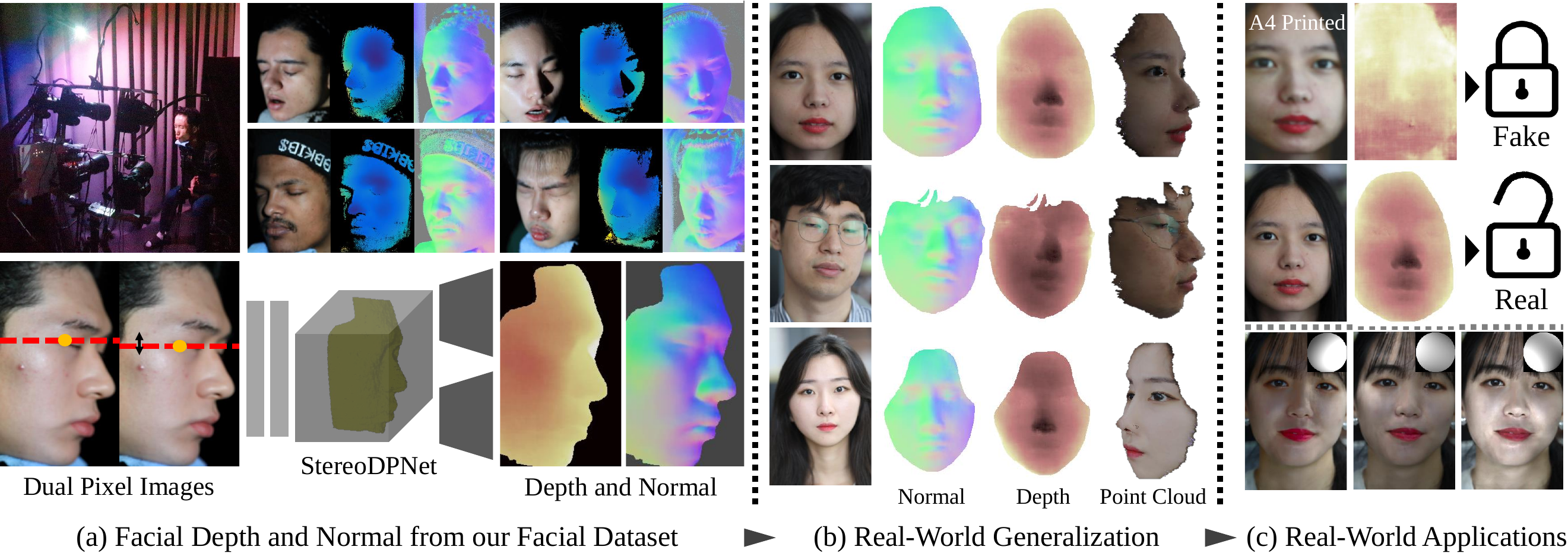}
\vspace{-2mm}
\caption{Using our Dual-Pixel Facial datasets, our network aims at generalized estimation of unmet facial geometry, which can be used for various applications, such as face spoofing or relighting.}
%
%
%
%
\label{fig:Teaser}
\vspace{-5mm}
\end{figure}


\vspace{-5mm}
\section{Related Work}
\label{sec:relatedwork}



\vspace{-2mm}
\noindent\textbf{Defocus-disparity in Dual-Pixel.}\quad 
Dual-Pixel images can be considered as a pair of stereo images since the DP camera captures two sub-aperture images with small parallax. However, since the DP camera is equipped with a micro-lens array in front of a camera sensor, the pixel disparity in DP images is extremely narrow (${-}4$px ${\sim}$ ${+}4$px)~\cite{du2net} compared to conventional stereo images. For this reason, Zhang~\etal~\cite{du2net} contend that a single pair of DP images is not suitable for cost volume-based disparity regression due to the narrow baseline~(${<}{\sim}1\text{mm}$). Other works~\cite{garg2019learning} also adopt simple 2D U-Net architectures for affine-transformed depth regression. 
Meanwhile, the disparity of DP images is induced by different left/right point spread functions (PSFs) instead of view parallax of stereo, called defocus-disparity~\cite{punnappurath2020modeling}. 
Based on this observation, Punnappurath~\etal~\cite{punnappurath2020modeling} propose an optimization-based disparity regression using a parametrized PSF.
The pioneering works~\cite{wadhwa2018synthetic,garg2019learning,punnappurath2020modeling} allow us to formulate depth-disparity conversion.

\noindent\textbf{Geometry dataset for Dual-Pixel.}\quad 
Owing to the growing research interest in DP photography, several real/synthetic DP datasets~\cite{garg2019learning,punnappurath2020modeling,DP-explore,realistic-DP} have been released. Garg~\etal~\cite{garg2019learning} propose a real-world DP dataset that includes scene-scale images captured by an array of smartphones. Despite of their success, the estimated depth is up to scale because their training data contains a relative-scaled 3D geometry computed by a multiview stereo algorithm (COLMAP~\cite{colmap}).
%
%
%
%
%
%

\noindent\textbf{Face dataset.}\quad 
Facial datasets have typically been created whenever new types of commercial imaging devices are introduced. Since it is only available to reconstruct faces from monocular images with a limited assumption~\cite{Wu_2020_CVPR}, many 3D face regression methods~\cite{richardson2017learning,feng2018prn,guo2020towards} rely on a given face morphable model~\cite{tran2018nonlinear,blanz1999morphable} and modify the shape with facial keypoints~\cite{feng2018prn,shang2020self} and landmarks~\cite{Deng_2019_CVPR_Workshops,bai2020deep}. Recently, several face regression models~\cite{wu2019mvf,bai2020deep} utilize multi-view images as input.
%
%
%
However, it is challenging to estimate facial geometry from DP images since blurry features are hardly captured from homogeneous regions of the face. This property brings difficulty in finding correspondence between left/right DP images.
We observe that the previous DP-oriented methods~\cite{garg2019learning,punnappurath2020modeling} have difficulty estimating the 3D geometry of human faces as well.
Many of the applications with facial images require both a high-quality depth and a surface normal for pleasing aesthetic effects~\cite{zhou2019deep}. Therefore, we satisfy the increasing industrial and academic demands by providing high-quality and absolute scale facial depth/normal maps that are captured with cameras with DP sensors.


\vspace{-3mm}
\section{Overview}
\label{sec:overview}
\vspace{-2mm}
This paper covers dual-pixel based facial understanding: from data acquisition (\Sref{sec:facial dataset}) to general estimation by stereoDPNet (\Sref{sec:architecture}). Different from natural images from typical cameras, dual-pixel sensors capture images having an extremely small range of disparity as well as defocus-blur, as shown in~\Fref{fig:dpcalibration}. Through our carefully designed dataset and network, we design a well-generalized methodology that even can infer facial geometry from unmet DP facial images.


\vspace{-2mm}
\section{Dual-Pixel Facial Dataset}
\label{sec:facial dataset}
\vspace{-2mm}
\begin{figure}[!t]
\centering
\includegraphics[width=0.98\linewidth]{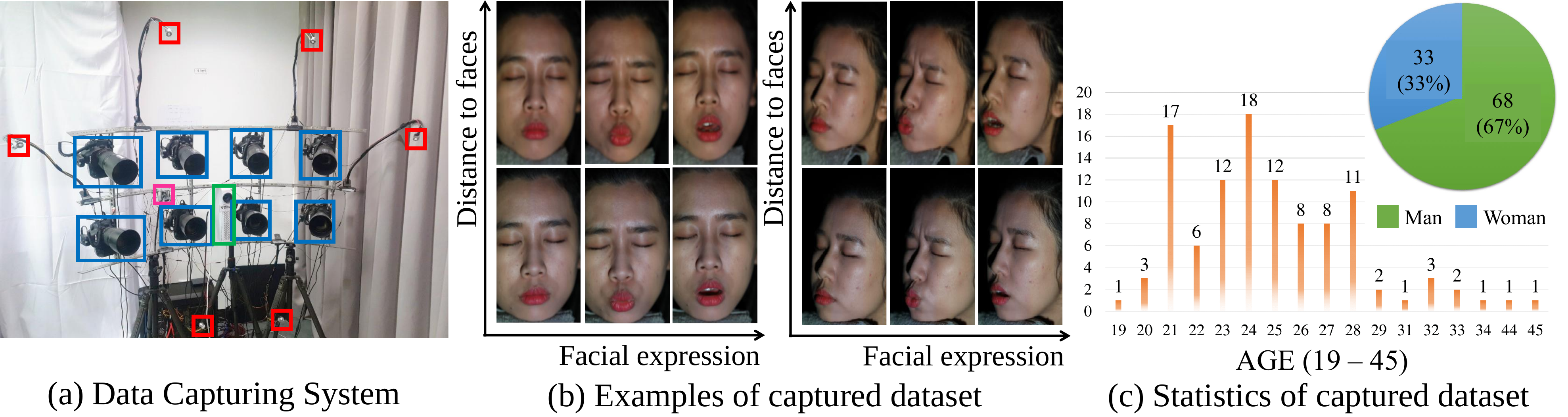}
\vspace{-2mm}
\caption{\textbf{Examples of our facial dataset.}
(a) The proposed hardware setup: ($2{\times}4$~multi-camera array~(\textcolor{blue}{blue}), $6$~LEDs~(\textcolor{red}{red}), a projector~(\textcolor{green}{green}), and a LED controller (\textcolor{magenta}{magenta}).
(b) DP images with various facial expressions (horizontal axis) and distances from the hardware to faces (vertical axis). There are two additional images taken in different heading directions (center/rightward).
(c) Our dataset has 68 men and 33 women and age distribution of them ranges from 19 to 45.
}
\label{fig:statistics}
\vspace{-5mm}
\end{figure}
In this section, we explain the construction of DP-oriented facial dataset. Since defocus-disparity in DP images is highly sensitive to image resolution, our dataset should contain both high resolution and high quality ground truth depth. By considering these requirements, we first explain the data configuration (\Sref{subsec:dataset configuration}), details of capturing system (\Sref{subsec:hardware_setup}), and describe a ground-truth depth/normal acquisition process (\Sref{subsec:ground truth acquisition}).


\vspace{-2mm}
\subsection{Dataset Configuration}
\label{subsec:dataset configuration}
Given an array of multiple DP cameras, we capture various human faces with different expressions and light conditions. The dataset consists of 135,744 photos, which are a combination of 101 people, eight cameras, seven different lighting condition, four facial heading directions (left, right, center and upward), three facial expressions (normal, open mouth and frown), and two fixed distances of subjects from the camera array, as illustrated in~\Fref{fig:statistics}.
The distances between the camera array and subjects range from 80~$cm$ to 110~$cm$. Since the focus distance is about 97~$cm$, our captured images contain both front focused and back focused cases.
Our dataset includes 44,352 female photos as well as 91,392 male photos, ages range from 19 to 45. The detailed statistics are provided in~\Fref{fig:statistics}-(c). In main experiments, we use 76 people ($76\%$) as a train set and the others ($24\%$) as a test/validation set without any overlap with the train set.

\vspace{-2mm}
\subsection{Hardware Setup}
\label{subsec:hardware_setup}
For facial data acquisition, we set up the DP-oriented camera-projector system. The system consists of eight synchronized Canon 5D Mark IV cameras on a 2$\times$4 grid, with one commercial projector ($1920\times 1080$ pixels) and six LED lights, as shown in~\Fref{fig:statistics} which enables capturing high-quality ground truths by using both structured light and photometric stereo. These cameras are available to capture DP images~\cite{abuolaim2020defocus,realistic-DP,DP-explore}. Each camera is equipped with a Canon 135$mm$ L lens. The $17^{\circ}$ field of view (FOV), which the lens affords, can cover an approximately $16.7cm \times 25 cm$ area at about one-meter distance, which is suitable for capturing human faces. We take our dataset with a camera aperture of F5.6, exposure time 1/30'', and ISO 1600.
The shape of the camera rig mimics a spherical dome with a one-meter radius. All of the cameras are located on the rig looking at the same point near the sphere's center. The projector is positioned at the center of the camera array. The LED lights are installed at various positions so that face images can be taken under varying lighting conditions.

\begin{figure}[!t]
\centering
\includegraphics[width=0.98\linewidth]{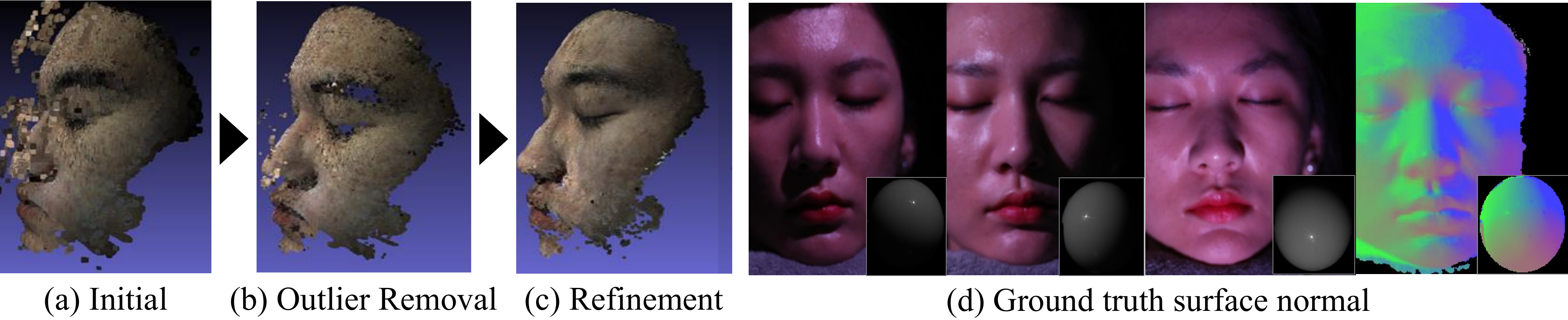}
\vspace{-2mm}
\caption{\textbf{Ground-truth depth and surface normal acquisition.}
(a) Initial depth from the structured light. 
(b) Depth after removing outliers. 
(c) Depth via fusion of the initial depth and the surface normal obtained from the photometric stereo in (c).}
\label{fig:gt_acquistion}
\vspace{-5mm}
\end{figure}
\vspace{-2mm}
\subsection{Ground Truth Data Acquisition}
\label{subsec:ground truth acquisition}
\vspace{-1mm}
Structured light systems are designed for high-quality 3D geometry acquisition under controlled environments by projecting pre-defined patterns on surfaces of objects~\cite{scharstein2003high,ha2015dense,chen2020auto} and by analyzing the projected patterns to measure 3D shapes of the objects. It is extensively used for ground-truth depth maps in stereo matching benchmarks~\cite{jensen2014large,aanaes2016large,silberman11indoor} and shape from shading~\cite{han2013high}. In this work, we tailor the structured light-based facial 3D reconstruction method~\cite{ha2015multi} with our well-synchronized multi-camera system. Thanks to our capturing system and structured light-based reconstruction method, we obtain dense, high-quality facial 3D corresponding to high-resolution DP images in~\Fref{fig:Teaser}-(a). Moreover, we calibrate point light directions by using a chrome ball and applying a photometric stereo in~\cite{nehab2005efficiently} to obtain accurate surface normal maps of subjects' faces in~\Fref{fig:gt_acquistion}(d). We utilize the RANSAC algorithm in obtaining both surface normal and albedo for robust estimation by excluding severe specular reflection. By using the surface normals, initial depth is refined by conforming the initial facial depth and the surface normal~\cite{nehab2005efficiently}, as illustrated in~\Fref{fig:gt_acquistion}(a), (b), and (c).

\begin{figure}[!t]
    \centering
    \includegraphics[width=0.99\linewidth]{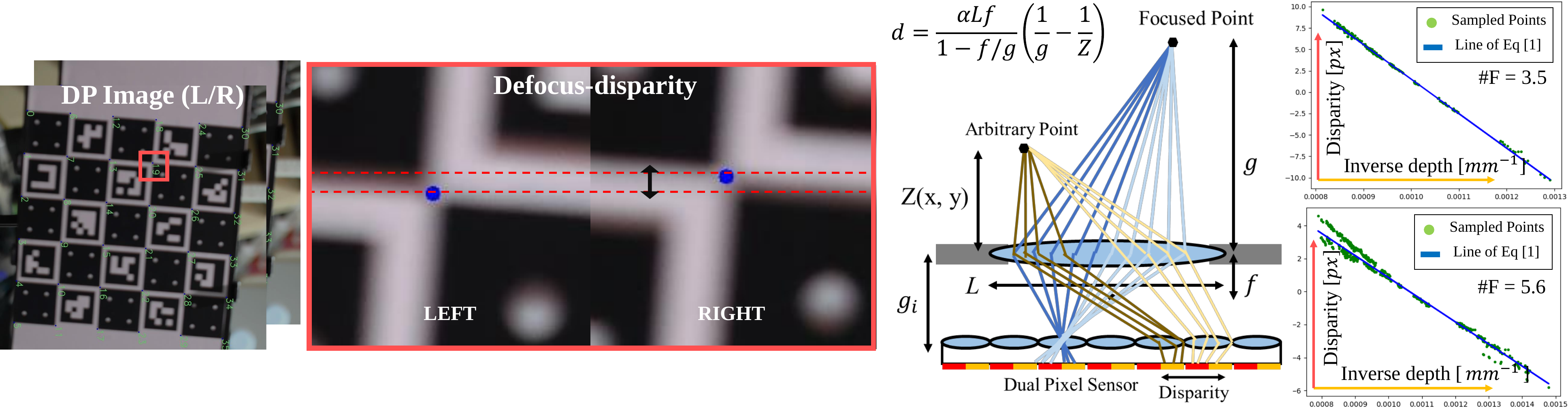}
    \vspace{-3mm}
    \caption{\textbf{Dual-Pixel Geometry.}
    Disparity in DP images exists in blurry regions, called defocus-disparity (left). The ground-truth depth obtained by a plane homography and a defocus-disparity from the matching pairs are used to robustly find parameters of~\Eref{eq:disparity_to_depth}. This relationship is used to back-project our prediction to metric-scale depth (right).}
    \label{fig:dpcalibration}
    \vspace{-5mm}
\end{figure}

To the end, we find an exact conversion between a defocus-disparity and a metric depth by using the relationship of signed defocus-blur~$\bar{b}(x,y)$ and disparity~$d$ (\Eref{eq:disparity_to_depth}) introduced in~\cite{garg2019learning} with the paraxial and thin-lens approximations.
\vspace{-2mm}
\begin{equation}
\begin{aligned}
\footnotesize
d(x, y) &=\alpha \bar{b}(x, y) \\
& \approx \alpha \frac{L f}{1-f / g}\left(\frac{1}{g}-\frac{1}{Z(x, y)}\right) \\
& \triangleq A(L, f, g)+\frac{B(L, f, g)}{Z(x, y)},
\end{aligned}
\label{eq:disparity_to_depth}
\vspace{-3mm}
\end{equation}
where $(x,y,Z(x,y))$ indicates 3D coordinates in camera space, $\alpha$ is a proportional term, and $L$ is the diameter of the camera aperture. $f$ represents the focal length of the lens and $g$ is the focus distance of the camera.
By using this relationship, we obtain a ground-truth defocus-disparity from the ground-truth depth in~\Fref{fig:dpcalibration}. This conversion is used to back-project our defocus-disparity to metric 3D space, and will be utilized to support absolute metric information for our depth and surface normal estimation network in~\Sref{sec:architecture}.

\vspace{-1mm}

\vspace{-2mm}
\section{Facial Depth and Normal Estimation}
\label{sec:architecture}
\vspace{-1mm}
Based on our dataset, we design StereoDPNet for the general estimation of facial depth and normal. In real applications, dual-pixel images can be captured with various camera parameters, such as focus distance or focal length. The proposed network should cope with various hardware setups as well. 

In this point of view, stereo matching methods~\cite{chang2018pyramid,xu2020aanet,khamis2018stereonet,zhang2019ga} show strength in generalization toward unmet environments and robust to camera configuration thanks to its corresponding search. Since a pair of DP images can also be regarded as stereo images having a short disparity range, we build our StereoDPNet based on the stereo-based depth/normal method~\cite{kusupati2020normal} for our dual-pixel based depth/normal estimation. 

\begin{figure*}[!t]
\centering
\includegraphics[width=0.99\linewidth]{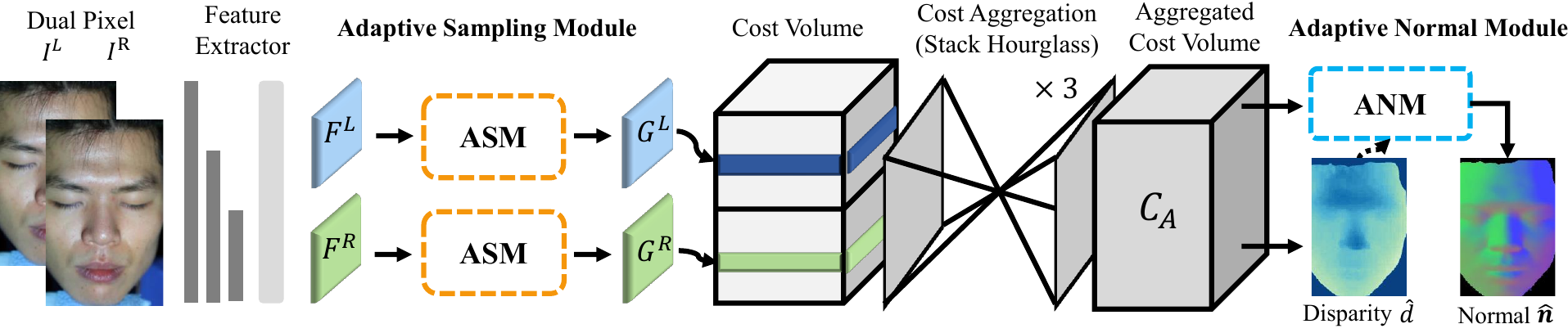}
\vspace{-3mm}
\caption{\textbf{Architecture of StereoDPNet.} Given DP images, our network is trained to infer facial depth/normal maps. Our two key modules, Adaptive Sampling Module and Adaptive Normal Module, overcome the extremely narrow baseline in DP images by capturing disparities in blurry regions. Note that we use the pre-defined relationship between disparity and depth in~\Sref{subsec:ground truth acquisition} to convert disparity to metric depth.}
\label{fig:StereoDPNet}
\vspace{-5mm}
\end{figure*}

Nonetheless, the stereo matching methods often require all-in-focus and well-textured images for pixel-wise correspondence search, which is not always guaranteed for DP images due to its defocus-blur and homogeneous/textureless faces. Thus, we propose our two novel modules, Adaptive Sampling Module (ASM) and Adaptive Normal Module (ANM), to further fit the properties of dual-pixel images. As in~\Fref{fig:StereoDPNet}, our StereoDPNet consists of four parts: feature extraction layer, ASM, cost aggregation layer, and ANM.

\vspace{-1mm}
\subsection{Overall Architecture}
\label{subsec:overall_architecture}
\vspace{-1mm}
Given DP images with left~$I^{L}$ and right $I^{R}$, stereoDPNet is trained to infer a disparity map $\hat{d}$ and a surface normal map $\hat{n}$. To do so, first, the feature extraction layer infers DP image features~$F^{L}$ and $F^{R}$, respectively.
%
%
Our feature extractor captures multi-scale information with a large receptive field by adopting Atrous Spatial Pyramid Pooling~\cite{chen2018encoder} and Feature Pyramid Network~\cite{lin2017feature} to encode various sizes of defocus blur in the DP images.
%
%
Second, using $F^{L}$ and $F^{R}$, the proposed ASM captures an amount of spatially varying blur in dynamic ranges, and then adaptively samples the features. Then, the sampled features $G^{L}$ and $G^{R}$ are stacked into a cost volume~$\mathcal{V}$. Third, the cost volume is aggregated through three stacked hourglass modules to infer the aggregated cost volume $C_{A}$. Lastly, this aggregated volume $C_{A}$ is used to regress a disparity map following the baseline and infer a surface normal map by ANM.
%
%
%
%
The details of ASM and ANM are described in~\Sref{subsec:adaptive_sampling_module} and~\Sref{subsec:adaptive_normal_module}.

\vspace{-3mm}
\subsection{Adaptive Sampling Module}
\label{subsec:adaptive_sampling_module}
\vspace{-1mm}

%

\begin{figure*}[!t]
\centering
\includegraphics[width=0.99\linewidth]{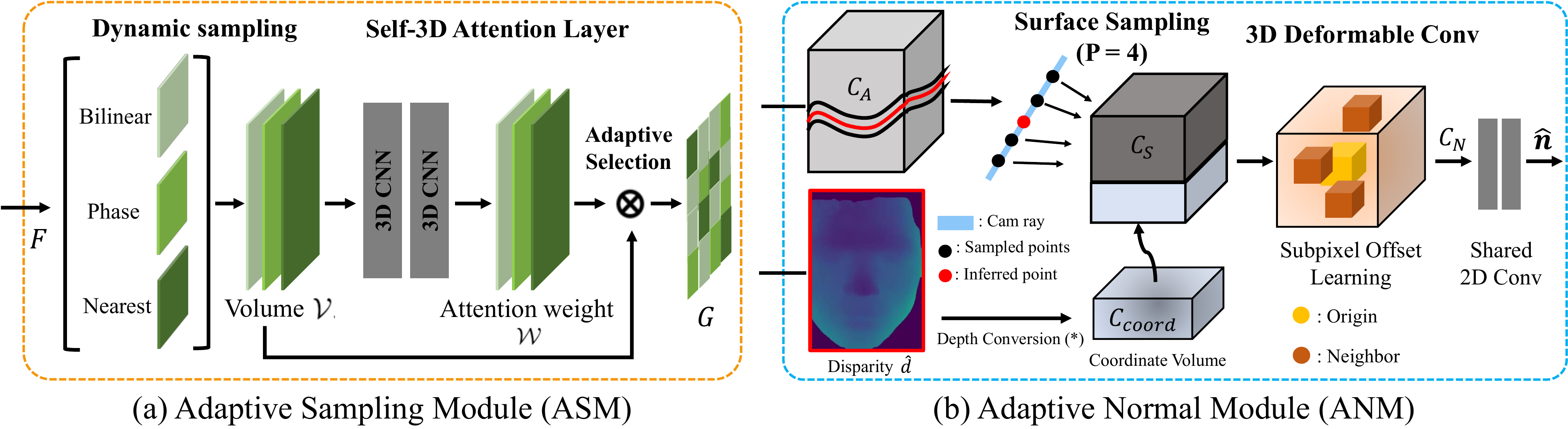}
\vspace{-3mm}
\caption{\textbf{Proposed Modules of StereoDPNet.} (a) Adaptive Sampling Module (ASM) consists of a dynamic sampling and a self-3D attention layer. (b) Adaptive Normal Module (ANM) consists of a surface sampling and a 3D deformable convolutional layer for surface normal regression.}
\label{fig:Modules}
\vspace{-6mm}
\end{figure*}

In contrast to the widely used stereo images such as KITTI Stereo Benchmark~\cite{geiger2012we,geiger2013vision}, dual-pixel images inherently have a small disparity range and defocus-blur. To cope with this issue, we design ASM inspired by defocus blur matching method~\cite{chen2015blur} and depth from narrow-baseline light-field image~\cite{jeon2015accurate}. As illustrated in~\Fref{fig:Modules}-(a), our ASM dynamically samples blurry texture features for narrow-baseline stereo matching.
To this end, the input features~($F^{L}$,$F^{R}$) pass through a \textit{dynamic feature sampling layer} and a \textit{self-3D attention layer} to obtain the locally dominant features $G^{L},G^{R}$.

According to Jeon~\etal~\cite{jeon2015accurate}, the sub-pixel shift from different sampling strategies to construct cost volume for matching provides varying results depending on the local scene configurations. In particular, phase-shift interpolation ensures a denser sampling field at sub-pixel precision and reduces the burden from blurriness compared to other interpolation methods~\cite{jeon2015accurate}. To take advantage of various conventional sampling methods, we incorporate them into ASM. The dynamic sampling layer in ASM is designed with a combination of nearest-neighbor, bilinear, and phase-shift interpolation, which can have various receptive fields to find varying blur sizes and can obtain subpixel-level shifted features. To this end, the shifted features from the three different sampling strategies are concatenated into one channel as a volumetric feature~$\mathcal{V}$.

To extract useful features from given volumetric feature~$\mathcal{V}$, we design a self-3D attention layer. Our self-3D attention layer adaptively selects sampling strategies to include prominent texture information in an extracted feature map. 
The layer consists of several 3D convolutional layers and the Sigmoid function.
This obtains a soft mask as attention map~$\mathcal{W}$ and selects features along the channel where $\mathcal{V}$ is concatenated. 
The soft mask $\mathcal{W}$ is multiplied with the feature $\mathcal{V}$ to sample useful features and the final feature volume $\mathcal{V}_{S}$ is produced through a softmax layer. Finally, the sampled features with the sub-pixel shift, $G^{L}$ and $G^{R}$, are obtained by averaging the volume $\mathcal{V}_{S}$.
The matching cost volume, constructed from the selected feature maps ($G^{L}, G^{R}$), contains rich texture information with relative blur and performs effective matching in homogeneous regions as well~\cite{chen2015blur}.

\vspace{-3mm}
\subsection{Adaptive Normal Module}
\label{subsec:adaptive_normal_module}
\vspace{-1mm}

As illustrated in~\Fref{fig:Modules}-(b), ANM aims to produce a surface normal map complementary to an estimated defocus-disparity map and to model 3D surface of human faces. The ANM consists of \textit{surface sampling module} to capture surface by sampling the aggregated cost volume~$C_{A}$ and \textit{deformable 3D convolutional layer} to consider dynamic ranges of neighbors to compute normal vectors.

According to~\cite{kusupati2020normal}, an accurately aggregated cost volume contains an implicit function representation of underlying surfaces for depth estimation. Since the surface normal mainly depends on the shape of the local surface, it is redundant to use all voxel embeddings in~$C_{A}$ for facial normal estimation. We thus sample the $P$ candidates of hypothesis planes among $M$ planes from the aggregated volume~$C_{A}$ using the estimated disparity map (\Eref{eq:disparity_regression}). Since the surface normal is defined with the metric scale depth, we convert disparity to a depth map using~\Eref{eq:disparity_to_depth} in~\Sref{subsec:ground truth acquisition} and provide this volumetric information with our network denoted as coordinate volume~$C_{coord}$. The details of surface sampling process is explained in the supplementary material.

Since a human face has a variety of curved local surfaces, we need to consider dynamic ranges of neighbors to extract a local surface from the sampled hypothesis planes $C_{S}$ in the previous stage. To do this, we follow the assumption of local plane in~\cite{long2021adaptive,qi2018geonet,long2020occlusion} and forms a local plane by a small set of neighbor points. Since these local patches have arbitrary shapes and sizes composed with its sampled neighboring points, we use 3D deformable convolutions~\cite{ying2020deformable} to consider the neighboring points within the dynamic ranges. The learnable offsets of the deformable convolution in 3D space allow us to adaptively sample neighbors and to find the best local plane. The final feature volume $C_{N}$ is predicted after passing two 3D deformable convolution layers to extract surface normal information from the sampled volume $C_{S}$.


\vspace{-2mm}
\subsection{Depth and Normal Estimation}
\label{subsec:depth_and_normal}
\vspace{-2mm}


The aggregated volume $C_{A}$ passes through a classifier to produce a final matching cost $\mathcal{A}$, and the softmax function $\sigma(\cdot)$ is applied to regress the defocus-disparity $\hat{d}$. Accordingly, we compute the disparity as follows:
\begin{equation}
\footnotesize
\hat{d}_{u,v}=\sum_{m=1}^{M} d^{m} \cdot \sigma\left(\mathcal{A}^{m}_{u,v}\right),
\label{eq:disparity_regression}
\end{equation}
where $\hat{d}_{u,v}$ is the defocus-disparity and $\mathcal{A}_{u,v}$ is the final matching cost at a pixel $(u,v)$. $M$ and $d^m$ are the range of defocus-disparity, and predefined discrete disparity labels, respectively, whose details are described in \Sref{subsec:implementation_detail}.
Following~\cite{chang2018pyramid}, we minimize a disparity loss $\mathcal{L}_{\text{disp}}$ using a smooth $L_1$ loss as follows:
\begin{equation}
\footnotesize
\mathcal{L}_{\text{disp}} = \frac{1}{H \cdot W} \sum_{u=1}^W\sum_{v=1}^H \mathcal{M}_{u,v} \cdot \operatorname{smooth}_{L_{1}}\left(d_{u,v}-\hat{d}_{u,v}\right),
\label{eq:smooth L1-loss}
\end{equation}
where $d_{u,v}$ is a ground-truth defocus-disparity at a pixel $(u,v)$ converted from the ground-truth metric scale depth and $\mathcal{M}_{u,v}$ is the facial mask in~\Sref{subsec:ground truth acquisition}.

For the surface normal estimation, shared 2D convolutions are applied to the feature volume $C_{N}$ to regress a surface normal. The final convolutional layers follow the same structure of the baseline architecture in~\cite{kusupati2020normal}. Finally, we train ANM by minimizing a cosine similarity normal loss $\mathcal{L}_{\text{normal}}$ as:
\begin{equation}
\footnotesize
\mathcal{L}_{\text{normal}} = \frac{1}{H \cdot W} \sum_{u=1}^W\sum_{v=1}^H \mathcal{M}_{u,v} \cdot \left(1-\mathbf{n}_{u,v} \cdot \hat{\mathbf{n}}_{u,v}\right),
\label{eq:cosine similarity loss}
\end{equation}
where $\mathbf{n}_{u,v}$ and $\hat{\mathbf{n}}_{u,v}$ are a ground-truth, and a predicted normal at a pixel $(u,v)$.
\begin{equation}
    \footnotesize
    \mathcal{L}_{\text{total}} = \mathcal{L}_{\text{disp}} + \mathcal{L}_{\text{normal}}
\end{equation}
Our StereoDPNet is fully supervised by our ground-truth depth/normal maps. The network is trained by minimizing the combination of \Eref{eq:smooth L1-loss} and \Eref{eq:cosine similarity loss}.

\vspace{-2mm}
\subsection{Implementation Details}
\label{subsec:implementation_detail}
The depth in our facial dataset ranges from 80~$cm$ to 110~$cm$ as described in~\Sref{subsec:dataset configuration} and the focus distance is about 97~$cm$. Therefore, our valid disparity range on original images is from -12 to 32 pixels. 
We note that the resolution of input images used is $1680 \times 1120$, which is downsampled four-fold due to GPU memory limitations.
We thus set the minimum and maximum disparity of~\Eref{eq:disparity_regression} to -4 and 12 pixels.
We also set the number of levels in the cost volume $M$ to 8, which represents a 0.5 pixel accuracy at least.
We train our network with a batch size of four, and use Adam optimizer~\cite{kingma2014adam} starting from the initial learning rate $10^{-4}$ with a constant decay of 0.5 at every 35 epochs.
\vspace{-2mm}

\begin{table*}[!t]
\centering
\scriptsize
\resizebox{1.00\linewidth}{!}{
\begin{tabular}{@{}cccccccc|ccc|cc@{}}

\\ \Xhline{4\arrayrulewidth}

\multirow{2}{*}{Method} & \multirow{2}{*}{Task} & & \multicolumn{5}{c}{Absolute error metric {[}mm{]} $\downarrow$} & \multicolumn{3}{c}{Affine error metric {[}px{]} $\downarrow$} & \multicolumn{2}{c}{Accuracy metric $\uparrow$} 

\\ \cline{4-13} 

& & & AbsRel & AbsDiff & SqRel & RMSE & RMSElog & WMAE & WRMSE & $1{-}\rho$ & ~$\delta {<} 1.01$~ & ~$\delta {<} 1.01^2$~

\\ \Xhline{2\arrayrulewidth}

 
PSMNet~\cite{chang2018pyramid} & ST & & 0.006 & 5.314 & 0.054 & 6.770 & 0.008 & 0.093 & 0.126 & 0.054 & 0.818 & 0.983 \\
StereoNet~\cite{khamis2018stereonet} & ST & & 0.005 & 4.306 & 0.038 & 5.811 & 0.006 & 0.112 & 0.150 & 0.087 & 0.903 & 0.991 
\\ 


DPNet~\cite{garg2019learning} & DP & & 0.008 & 7.175 & 0.092 & 8.833 & 0.010 & 0.110 & 0.148 & 0.086 & 0.688 & 0.959 \\
MDD~\cite{punnappurath2020modeling} & DP & & - & - & - & - & - & 1.830 & 2.348 & 0.575 & - & - 

\\ 

BTS~\cite{lee2019big} & M & & 0.007 & 6.575 & 0.081 & 8.102 & 0.009 & 0.111 & 0.150 & 0.077 & 0.731 & 0.964 
\\ 


NNet~\cite{kusupati2020normal} & DN & & \second{0.004} & \second{3.608} & \second{0.027} & \second{4.858} & \second{0.005} & \second{0.073} & \second{0.102} & \second{0.048} & \second{0.934} & \second{0.995} 

\\ \Xhline{2\arrayrulewidth}

\textbf{Ours} & DN & & \first{0.003} & \first{2.864} & \first{0.019} & \first{3.899} & \first{0.004} & \first{0.064} & \first{0.091} & \first{0.034} & \first{0.966} & \first{0.995}

\\ \Xhline{4\arrayrulewidth}

\end{tabular}
}
\vspace{1mm}
\caption{\textbf{Depth Benchmark Results.} We show that our proposed method outperforms the existing stereo matching methods (PSMNet~\cite{chang2018pyramid}, StereoNet~\cite{khamis2018stereonet}), DP-oriented state-of-the-art methods (DPNet~\cite{garg2019learning}, MDD~\cite{punnappurath2020modeling}), monocular depth estimation (BTS~\cite{lee2019big}), and depth/normal network for stereo matching (NNet~\cite{kusupati2020normal}). 
Note that since MDD adopts another a defocus-disparity geometry different from ~\cite{garg2019learning}, it is not measured by the absolute metrics.
ST, DP, M, and DN denotes ``Stereo Matching", ``DP-oriented method", ``Monocular", and ``Depth and Normal", respectively.
}
\label{tab:depth_benchmark}
\vspace{-7mm}
\end{table*}
\begin{table*}[!t]
\centering
\scriptsize
\setlength{\tabcolsep}{1pt}
\begin{tabular}{ccc|cc|ccc|cc|cc}

\\ \Xhline{4\arrayrulewidth}

\multirow{2}{*}{Method} & \multicolumn{2}{c}{ANM} & \multicolumn{2}{c}{Absolute {[}mm{]} $\downarrow$} & \multicolumn{3}{c}{Affine {[}px{]} $\downarrow$} & \multicolumn{2}{c}{Accuracy $\uparrow$} & \multicolumn{2}{c}{Normal {[}deg{]} $\downarrow$} 
\\ \cline{2-12}

& ~~SS & D3D & AbsDiff & RMSE & WMAE & WRMSE & $1{-}\rho$ & ~$\delta {<} 1.01$~ & ~$\delta {<} 1.01^{2}$~ & MAE & RMSE 
\\ \Xhline{2\arrayrulewidth}

ASM Only & & & 4.895 & 6.223 & 0.095 & 0.127 & 0.056 & 0.850 & 0.992 & - & - \\
NNet~\cite{kusupati2020normal} & & & 3.608 & 4.858 & 0.073 & 0.102 & 0.048 & 0.934 & \second{0.995} & 9.634 & 11.877 \\
ASM + NNet & & & 3.271 & 4.434 & \second{0.064} & \second{0.090} & \first{0.033} & 0.947 & \first{0.997} & 9.072 & 11.045 \\
ASM + NNet & \cmark & & 3.214 & 4.519 & \first{0.062} & \first{0.089} & 0.037 & 0.943 & 0.990 & \second{8.894} & \second{10.837} 


\\ \Xhline{2\arrayrulewidth}

\textbf{StereoDPNet} & \cmark & \cmark & \first{2.864} & \first{3.899} & \second{0.064} & 0.091 & \second{0.034} & \first{0.966} & \second{0.995} & \first{7.479} & \first{9.386} 

\\ \Xhline{4\arrayrulewidth}
\end{tabular}
\vspace{1mm}
\caption{\textbf{Ablation Study of ANM.} NNet~\cite{kusupati2020normal} is a baseline model of our overall architecture. We compare the performance of depth and surface normal estimation by adding each component. SS denotes ``Surface Sampling" and D3D denotes ``Deformable 3D convolution" of ANM respectively.}
\label{tab:ablation_normal}
\vspace{-8mm}
\end{table*}
\begin{table}[!t]
\centering
\scriptsize
\setlength{\tabcolsep}{2.0pt}
\renewcommand{\arraystretch}{1.1}
\begin{tabular}{lccc|ccc}

\\ \Xhline{4\arrayrulewidth}

\multicolumn{1}{c}{\multirow{2}{*}{Method}} & \multicolumn{3}{c}{Absolute error metric {[}mm{]} $\downarrow$} & \multicolumn{2}{c}{Accuracy metric $\uparrow$}
\\ \cline{2-6}
& ~~AbsDiff~ & ~SqRel~ & ~RMSE~ & ~$\delta {<} 1.01$~ & ~$\delta {<} 1.01^2$~

\\ \Xhline{2\arrayrulewidth}

Bilinear (Bi) & 7.956 & 0.116 & 9.842 & 0.615 & 0.928 \\
Phase & 8.287 & 0.132 & 10.487 & 0.606 & 0.916 \\
Phase + Bi & 6.030 & 0.067 & 7.522 & 0.754 & 0.980 \\
Nearest + Bi & 5.841 & \second{0.062} & 7.287 & 0.772 & 0.984 \\
Nearest + Phase & \second{5.831} & \second{0.062} & \second{7.247} & \second{0.773} & \second{0.985}
\\ \hline
\textbf{ASM} & \first{4.895} & \first{0.045} & \first{6.223} & \first{0.850} & \first{0.992} 

\\ \Xhline{4\arrayrulewidth}

\end{tabular}
\vspace{2mm}
\caption{\textbf{Ablation Study of ASM.} We test various sampling strategies in ASM and determine the final structure of ASM. Here, we only use ASM to strictly compare the inference of each sampling strategies. Bi denotes bilinear sampling.}
\label{tab:ablation_sampling}
\vspace{-7mm}
\end{table}


\begin{figure*}[!t]
\centering
\includegraphics[width=0.98\linewidth]{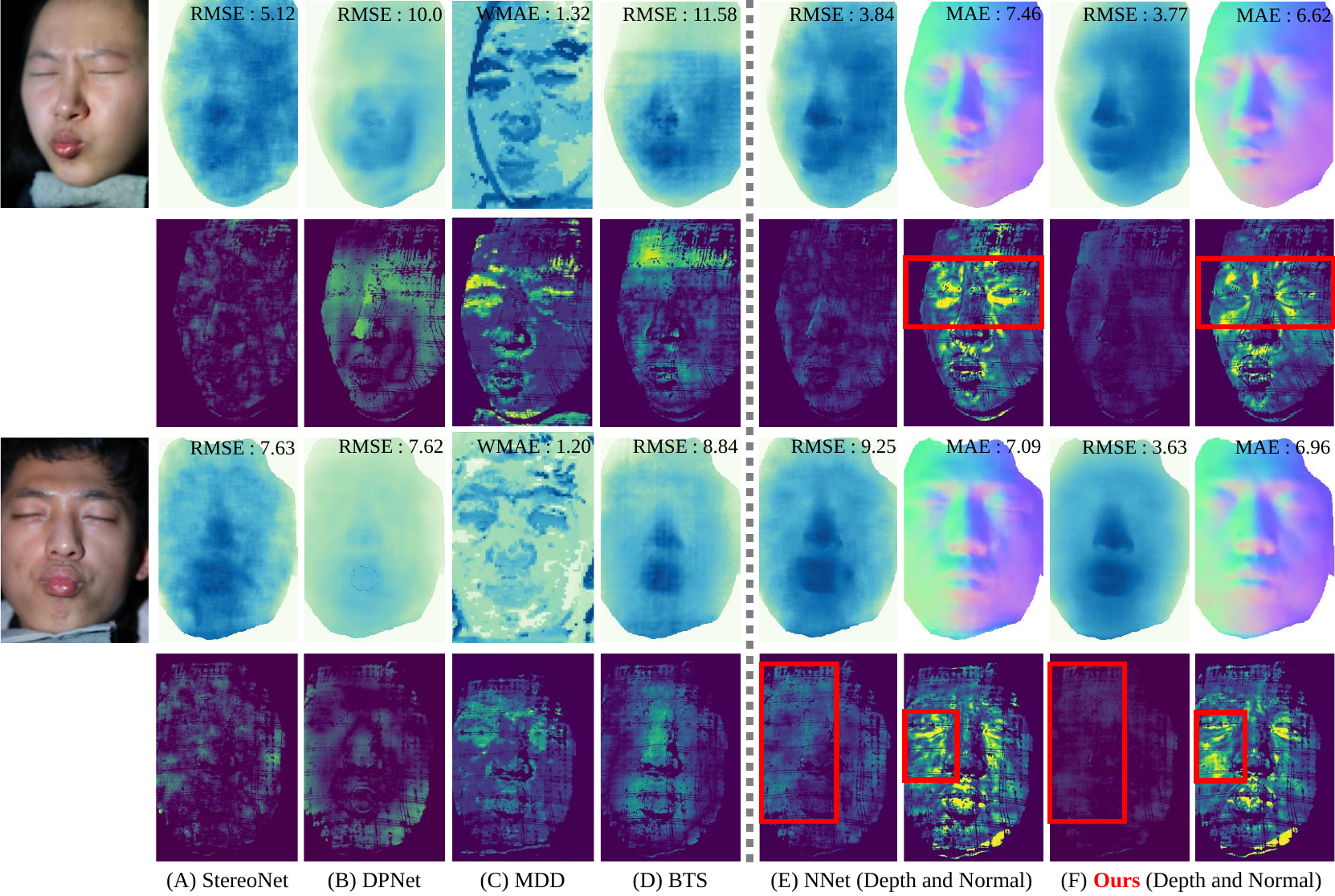}
\vspace{-3mm}
\caption{\textbf{Qualitative results on test set.} We report AbsRel map of depths and MAE map of normals as the error map of predictions. The error map from MDD is the WMAE map because it predicts relative scale depth maps. We note that the range of error map is 0.0 $\sim$ 1.0 (AbsRel [mm]) and 0.0 $\sim$ 15.0 (MAE [degree]).}
\vspace{-3mm}
\label{fig:qualitative_result}
\end{figure*}
\begin{figure*}[!t]
\centering
\includegraphics[width=0.98\linewidth]{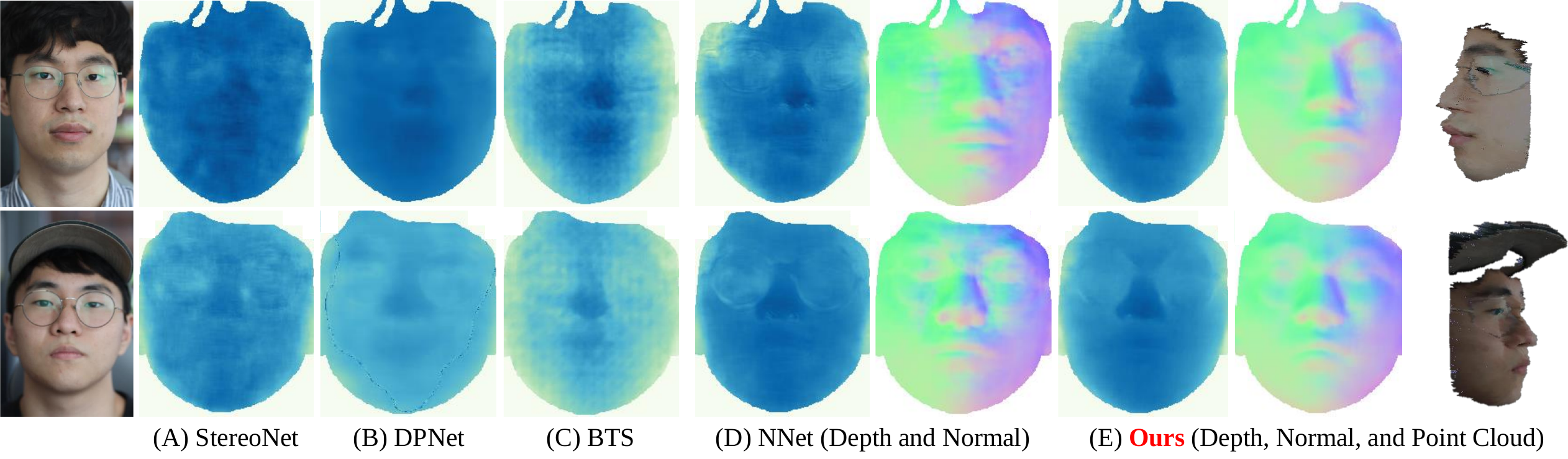}
\vspace{-3mm}
\caption{\textbf{Real-world results.} We capture faces in unmet real-world and compare our method with the others in~\Tref{tab:depth_benchmark}. StereoDPNet clearly captures surface and boundary depth of the face. Please refer to the supplementary material for more examples.}
\vspace{-5mm}
\label{fig:real_world_result}
\end{figure*}

\section{Experiments}
\label{sec:experiments}
\vspace{-2mm}
To evaluate the effectiveness and the robustness of our work, we carry out various experiments on our dataset as well as DP images captured under real-world environments.
For a fair comparison, all the methods are trained on identical training sets of our dataset from scratch. We then evaluate the quality of estimated depth/normal maps on the same test split of our facial dataset. 
Note that we use the facial mask for training and the test, given by the data acquisition process in~\Sref{subsec:dataset configuration}. 
For the real-world samples, we use a facial mask from a pretrained face segmentation network\footnote{\url{https://github.com/zllrunning/face-parsing.PyTorch}}.

\vspace{-1mm}
\subsection{Comparison Results}
\label{subsec:comparison}
\vspace{-1mm}
\noindent\textbf{Evaluation Metrics.}\quad
In our dataset benchmark, we convert our predicted disparity to depth (\Sref{subsec:ground truth acquisition}). Thus, we use both evaluation metrics in a public benchmark suite\footnote{\url{http://www.cvlibs.net/datasets/kitti/}}: AbsRel, AbsDiff, SqRel, RMSE, RMSElog, and inlier pixel ratios ($\delta$ $<$ 1.01$^i$ where $i \in$ $\{1, 2, 3\}$)\footnote{All equations of the metrics are described in Supplementary material.} and affine invariant metrics~\cite{garg2019learning} for the evaluation of predicted disparity/depth. To measure the quality of a surface normal map, we utilize a Mean Angular Error (MAE) and a Root Mean Square Angular Error (RMSAE) in degree unit following the DiLiGenT benchmark~\cite{shi2016benchmark}.

\noindent\textbf{Depth Benchmark.}\quad
We compare our method with recent DP-based depth estimation approaches~\cite{garg2019learning,punnappurath2020modeling} as well as widely used stereo matching networks~\cite{chang2018pyramid,khamis2018stereonet}, a depth/normal network for stereo matching, NNet~\cite{kusupati2020normal} and a state-of-the-art monocular depth estimation network, BTS~\cite{lee2019big}, whose results are reported in~\Tref{tab:depth_benchmark} and in~\Fref{fig:qualitative_result}. 
Since there is no published code for~\cite{garg2019learning}, we implement DPNet~\cite{garg2019learning} to predict disparity instead of inverse depth following them, and check that the performance is similar on their dataset. 

Due to the small range of the defocus-disparity from DP images (-4px to 12px), the cost volume with the discrete hypotheses leads to unstable training~\cite{du2net}.
As a result, typical stereo matching based depth estimations, PSMNet~\cite{chang2018pyramid}, StereoNet~\cite{khamis2018stereonet} do not work well. The methods in~\Tref{tab:depth_benchmark} except ours fail to handle defocus blur or struggle to find correspondances in human faces.
For example, both DPNet and BTS suffer from blurry predictions, and MDD~\cite{punnappurath2020modeling} is sensitive to the textureless regions in human faces.
Although NNet~\cite{kusupati2020normal} show relatively promising results, our method still outperforms them.
Moreover, real-world results in~\Fref{fig:real_world_result} demonstrates that our network is specialized in finding defocus-disparity from facial DP images and robust to blur which produces robust results from the unmet facial scene.


\begin{figure}[!t]
\centering
\includegraphics[width=0.95\linewidth]{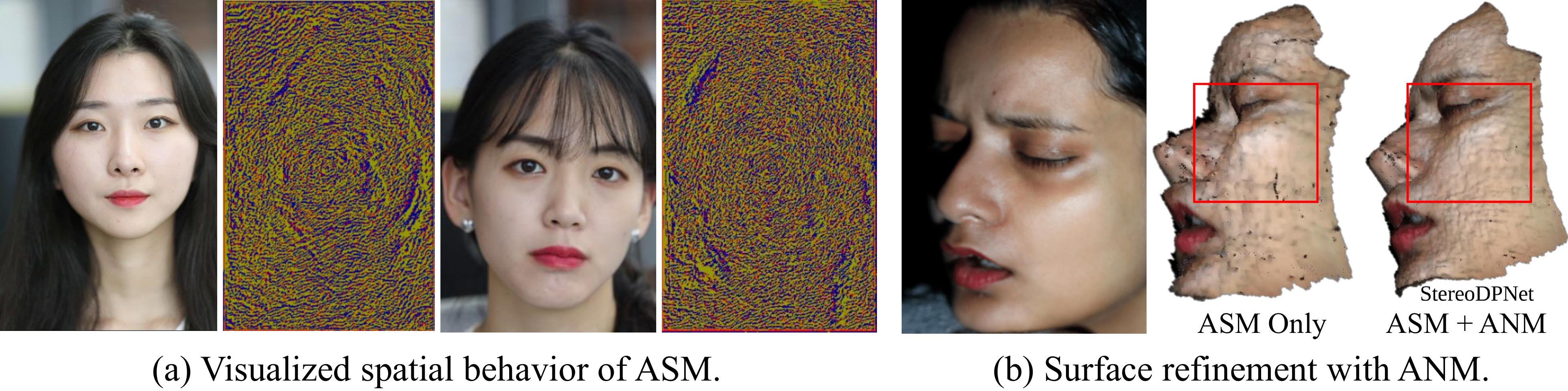}
\vspace{-3mm}
\caption{(a) Spatial behavior of ASM (\textcolor{darkyellow}{bilinear}, \textcolor{blue}{nearest}, and \textcolor{red}{phase}). (b) We compare 3D point cloud of the method of only ASM and with our full method of ASM and ANM. The result demonstrates that ANM refines surface of the facial 3D.}
\label{fig:anm_analysis}
\vspace{-5mm}
\end{figure}

\noindent\textbf{Surface Normal Benchmark.}\quad
To the best of our knowledge, this is the first attempt to estimate both the surface normal and the defocus-disparity from single DP images. Since the basic structure of ANM is derived from the recent depth and normal network~\cite{kusupati2020normal} for multi-view stereo, we show the performance improvement of our ANM, compared to the baseline method~\cite{kusupati2020normal} by adding each component in~\Tref{tab:ablation_normal}. We find that joint learning of disparity and surface normal leads to geometrically consistent and high-quality depth and surface normal shown in~\Fref{fig:anm_analysis}, which has been demonstrated in previous works~\cite{Qiu_2019_CVPR,im2015high}.

\vspace{-2mm}
\subsection{Ablation Study}
\label{subsec:ablation}
\vspace{-1mm}
\noindent\textbf{Analysis of ASM and ANM.}\quad
First, we compare various subpixel sampling strategies in ASM as an ablation study.
In~\Tref{tab:ablation_sampling}, including whole attention maps from three different interpolations as proposed in ASM, shows the best performance over any combination of two interpolations. We also provide spatial attention of $\mathcal{W}$ in ASM following the illustration scheme for selective matching costs in~\cite{jeon2018depth}. It shows that different sampling schemes are adaptively chosen.
Second, we show that our surface normal estimation greatly improves the prediction of the disparity in~\Tref{tab:ablation_normal}. ANM captures local surface and refines the surface via cost volume and leads to major performance improvement as shown in~\Fref{fig:anm_analysis}. We attribute this outstanding performance to our sub-modules (ASM and ANM), which overcome the extremely narrow baselines in DP images.

\begin{figure}[!t]
\centering
\includegraphics[width=0.96\linewidth]{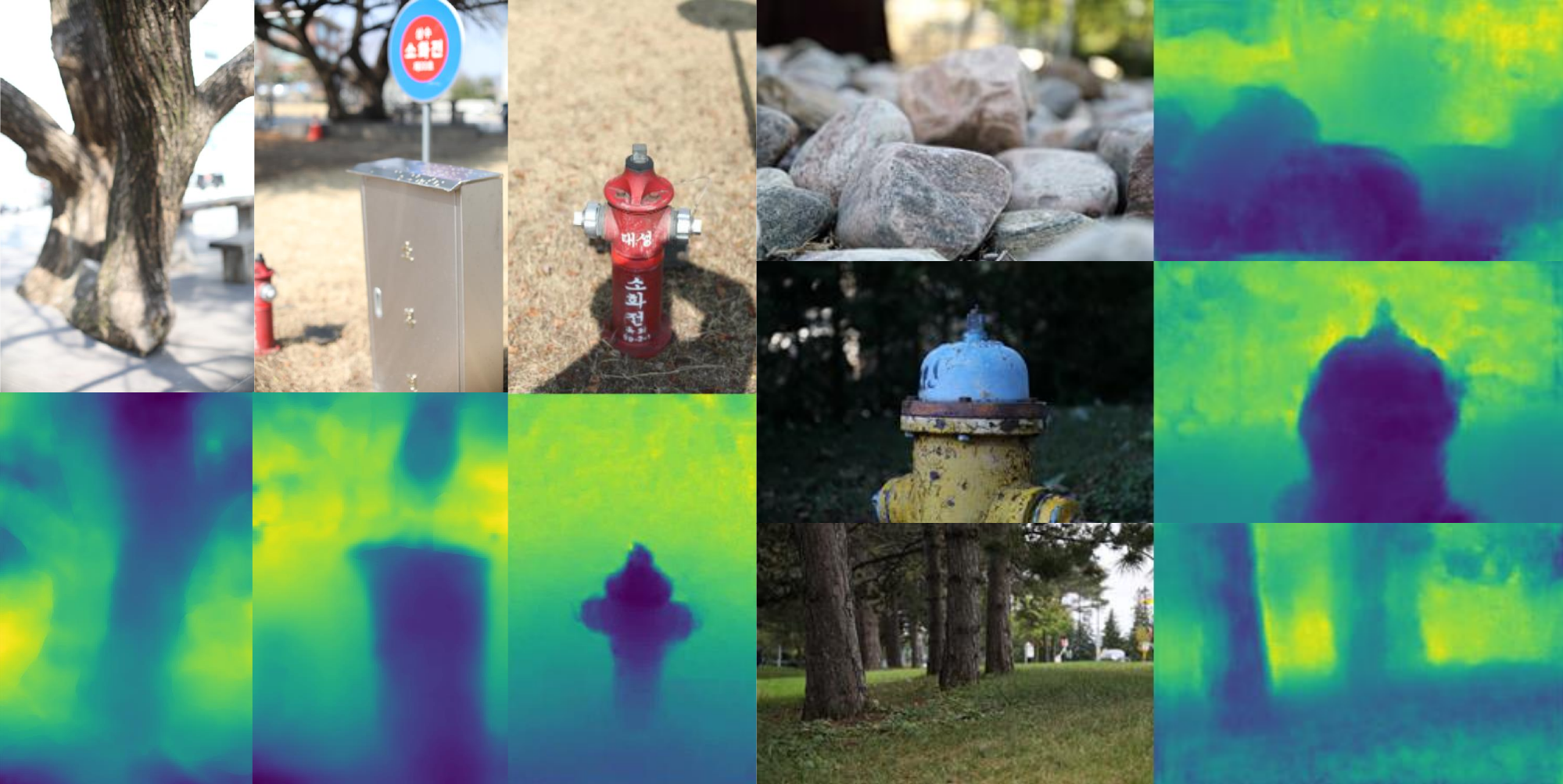}
\vspace{-3mm}
\caption{\textbf{Depth from single DP images in the wild.}
We show depth estimation results of StereoDPNet on outdoor photos, which are directly captured by us (left side) and in a public real-world DP dataset~\cite{abuolaim2020defocus} for deblurring (right side).}
\label{fig:general_scene}
\vspace{-3mm}
\end{figure}

\begin{figure}[t]
\centering
\includegraphics[width=0.97\linewidth]{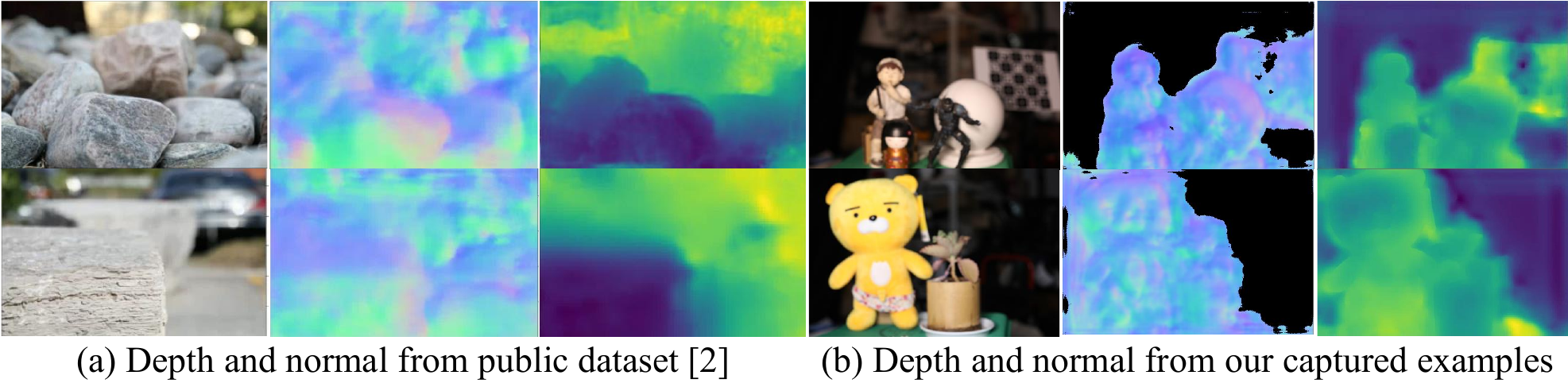}
\vspace{-3mm}
\caption{
\textbf{Depth and normal results in the wild.} (a) public DP dataset~\cite{abuolaim2020defocus}, and (b) our captured natural DP images. 
}
\label{fig:general_scene2}
\vspace{-6mm}
\end{figure}

\noindent\textbf{Real World Experiment.}\quad
To verify the generalizability of our method, we newly capture outdoor DP images using the Canon DSLR camera. We capture faces with various camera parameters (focus distance from 1.0m to 1.5m and F-number from 2.0 to 7.1) to demonstrate our method's robustness. Some of the results are shown in~\Fref{fig:real_world_result}.
Surprisingly, our StereoDPNet trained solely on our facial dataset also works well with general scenes (\Fref{fig:general_scene}, \Fref{fig:general_scene2}), which demonstrates that our method is generalized well. Some of the scenes in~\Fref{fig:general_scene} are from public DP dataset~\cite{abuolaim2020defocus} (focus distance from 1.46m to 1.59m).


\begin{figure}[!t]
\centering
\includegraphics[width=0.97\linewidth]{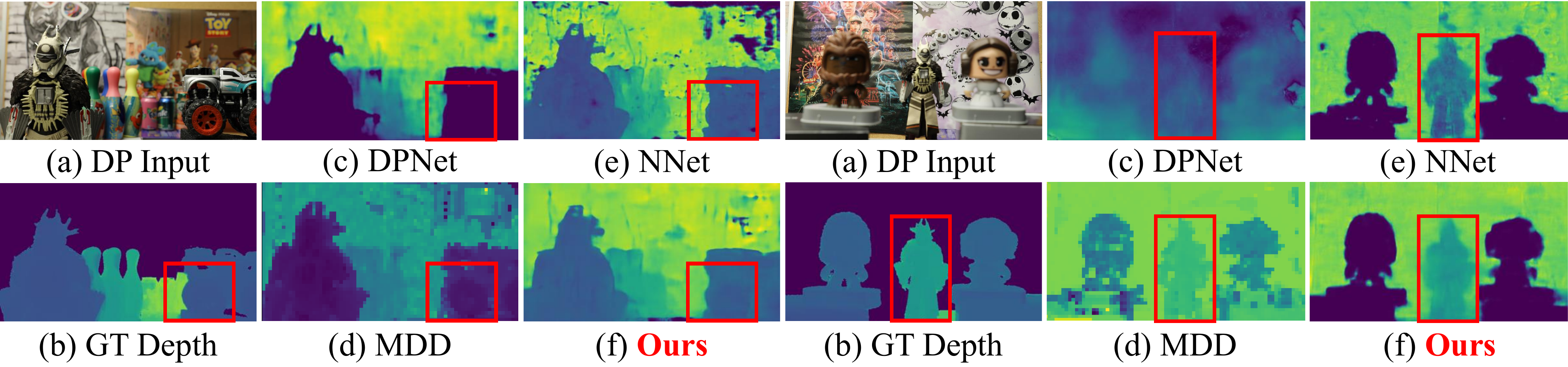}
\vspace{-3mm}
\caption{\textbf{Generalization  on  the  public  dataset~\cite{punnappurath2020modeling}.}
We show qualitative results of depth from the methods in~\Tref{tab:depth_benchmark} on the public dataset~\cite{punnappurath2020modeling}.}
\label{fig:more_results}
\vspace{-5mm}
\end{figure}

\noindent\textbf{Generalization on the public dataset~\cite{punnappurath2020modeling}.}\quad
We conduct an additional experiment on another real-world DP dataset~\cite{punnappurath2020modeling} to validate the generalization of our network. Although our method is generalized well in real-world scenarios in~\Fref{fig:general_scene} and in~\Fref{fig:general_scene2}, we augment our network by using an additional synthetic DP dataset in~\cite{realistic-DP}. This is because there is no currently available large real-world DP dataset captured with DSLR for training except ours. For a fair comparison, we don't apply any post-processing (\ie bilateral or guided filter) to the predictions. As illustrated in \Fref{fig:more_results} and \Tref{tab:extensive}, our network shows promising results on the non-facial dataset~\cite{punnappurath2020modeling} as well.

\begin{table}[t]
\centering
\scriptsize
\resizebox{0.90\linewidth}{!}{
\begin{tabular}{ccccccc}

\\ \Xhline{4\arrayrulewidth}

\multicolumn{1}{c|}{\multirow{2}{*}{Metrics}} & \multicolumn{6}{c}{Method} 
\\ 

\multicolumn{1}{c|}{} & 
{PSMNet~\cite{chang2018pyramid}} & {StereoNet~\cite{khamis2018stereonet}} & {DPNet~\cite{garg2019learning}} & {MDD~\cite{punnappurath2020modeling}} & {NNet~\cite{kusupati2020normal}} & \textbf{Ours}     

\\ \Xhline{2\arrayrulewidth}

\multicolumn{1}{c|}{WMAE ($\downarrow$)} & \second{0.102} & 0.111  & 0.132 & 0.107 & 0.103 & \first{0.085} \\
\multicolumn{1}{c|}{WRMSE ($\downarrow$)} & 0.154  & 0.214  & 0.192 & 0.168 & \second{0.143}  & \first{0.133}  \\
\multicolumn{1}{c|}{$1{-}\rho$ ($\downarrow$)} & 0.351 & \second{0.261}  & 0.420 & \first{0.187} & 0.345 & 0.276

\\ \Xhline{4\arrayrulewidth}

\end{tabular}
}
\vspace{2mm}
\caption{\textbf{Comparisons on the public dataset~\cite{punnappurath2020modeling}.} We provide quantitative comparison result of the methods in~\Tref{tab:depth_benchmark} on the public dataset~\cite{punnappurath2020modeling}.} 
\vspace{-5mm}
\label{tab:extensive}
\end{table}

\vspace{-2mm}
\section{Conclusion}
\label{sec:conclusion}
\vspace{-2mm}
We present a high-quality facial DP dataset incorporating 135,744 face images for 101 subjects with corresponding depth maps in metric scale and surface normal maps.
Moreover, we introduce DP-oriented StereoDPNet for both depth and surface normal estimation. StereoDPNet successfully shows impressive results in the wild by effectively handling the narrow baseline problem in DP.

\noindent\textbf{Potential societal impact.}\quad
We have already received consent from participants to use our facial dataset for only academic purposes. Thus, our dataset will be available to the computer vision community to promote relevant research.

\noindent\textbf{Limitation.}\quad
Although we show that our method is generalized to real-world DP scenes with various focus distances, our dataset is captured with fixed focus distance which is a clear limitation. Moreover, our ground truth acquisition of surface normal doesn't fully consider the complex specular reflection of the face which still remains as a challenging issue~\cite{lattas2020avatarme,song2020recovering}. We also recognize that our dataset has an inherent bias in skin tone. However, these limitations can be resolved by capturing more dataset with various camera parameters and considering advanced non-Lambertian shape from shading methods~\cite{lichy2021shape,boss2020two,lattas2020avatarme}. We will refine the dataset sustainably and try to resolve these limitations.

\noindent\textbf{Acknowledgements.}\quad
This work is in part supported by the Ministry of Trade, Industry and Energy (MOTIE) and Korea Institute for Advancement of Technology (KIAT) through the International Cooperative R\&D program in part (P0019797), `Project for Science and Technology Opens the Future of the Region' program through the INNOPOLIS FOUNDATION funded by Ministry of Science and ICT (Project Number: 2022-DD-UP-0312), and also supported by the Samsung Electronics Co., Ltd (Project Number: G01210570).

\appendix
\newpage

\section*{Supplementary material}
\label{sec:supple_overview}

This is a supplementary material for the paper, \textit{Facial Depth and Normal Estimation using Single Dual-Pixel Camera}. We will further describe details: ground truth acquisition process of our facial DP dataset (\Sref{sec:supple_gt_acquisition}), concrete description of evaluation metrics (\Sref{sec:supple_metrics}), precise algorithm of our Adaptive Normal Module (\Sref{sec:supple_algorithm}), and generalization experiments using our network and dataset (\Sref{sec:supple_results}).

\section{Ground-Truth Acquisition}
\label{sec:supple_gt_acquisition}

In this section, we describe the precise ways of ground truth depth acquisition processes from our hardware setup. These processes consist of three sub-processes: (1)~depth from structured light~(\Sref{subsec:supple_structuredlight}) 
(2)~multi-view depth refinement~(\Sref{subsec:multi-view_refinement}) 
(3)~normal-assisted depth refinement~(\Sref{subsec:Normal-guided Depth Refinement}). To this end, we explain detail of our process to acquire parameters of~\Eref{eq:disparity_to_depth} in manuscript.

\subsection{Depth from Structured Light}
\label{subsec:supple_structuredlight}
\begin{figure}[!t]
\centering
\includegraphics[width=0.99\linewidth]{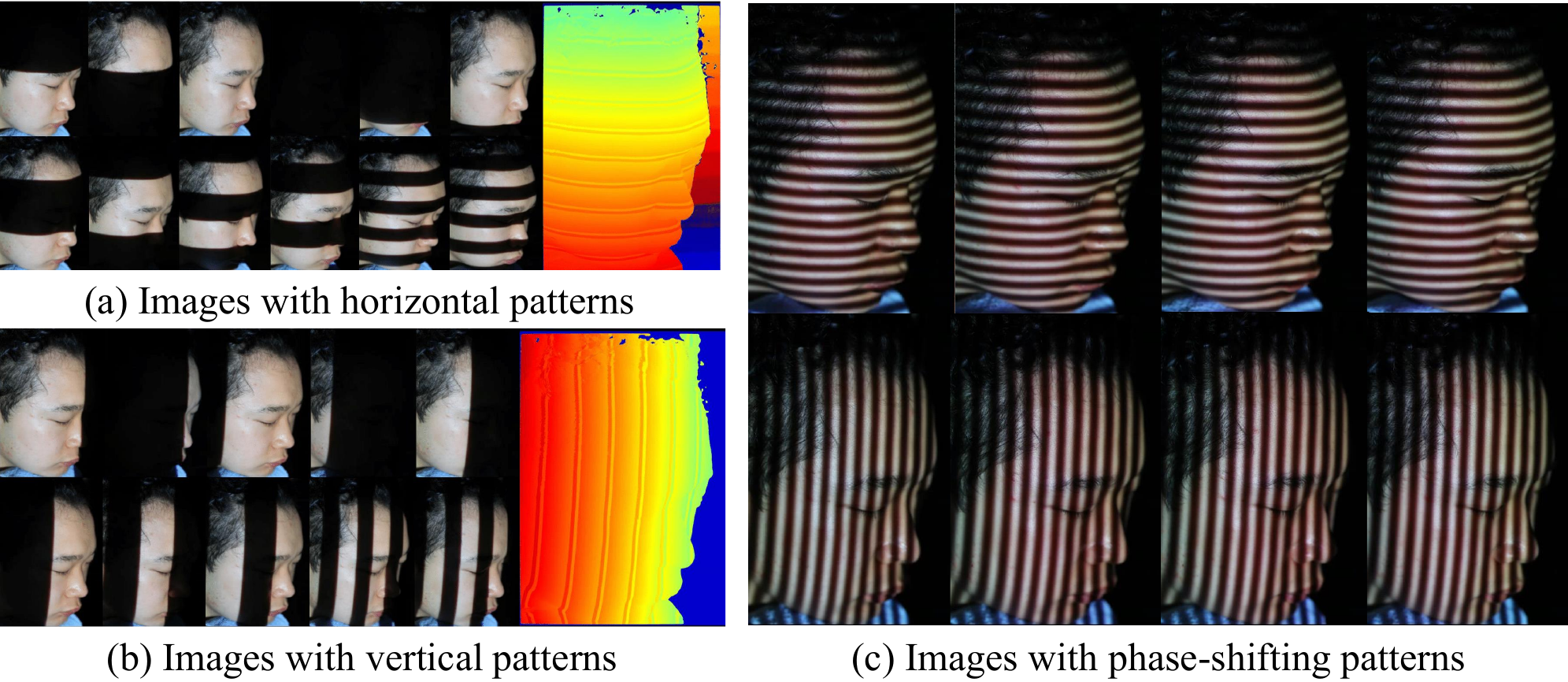}
\vspace{-2mm}
\caption{\textbf{Gray-code patterns and decoding process used in Structured-Light based ground-truth acquisition.} (a) Images captured with 12 horizontal patterns and acquired unwrapped phase image. (b) Images captured with 10 vertical patterns and acquired unwrapped phase image. (c) Images captured with phase-shifting patterns to refine acquired unwrapped phase images.}
\label{supplefig:structured_fig1}
\vspace{-2mm}
\end{figure}

Under the carefully designed hardware setup, we start to acquire initial ground truth depth maps from structured light. 
We would like to briefly explain detail processes to acquire unwrapped phase images from coded patterns.
As in~\cite{ha2015dense,ha2015multi}, we use $12$ horizontal patterns and $10$ vertical patterns consisting of $6$-bit inverse gray-code~(\Fref{supplefig:structured_fig1}-(a),(b)) to get unwrapped phase images, and $8$ phase-shifting patterns~\cite{geng2011structured,salvi2010state}~(\Fref{supplefig:structured_fig1}-(c)) for phase correction~\cite{je2015color}. These pattern images are used to acquire two (horizontal/vertical) unwrapped phase images per camera view.
Then, we estimate dense correspondences based on a standard phase unwrapping method~\cite{salvi2010state}. After that, consistency between multi-view unwrapped phase images' intensity is used to get high quality of facial depth following~\cite{ha2015multi}.

\subsection{Multi-View Depth Refinement}
\label{subsec:multi-view_refinement}
\begin{figure}[!t]
\centering
\includegraphics[width=0.82\linewidth]{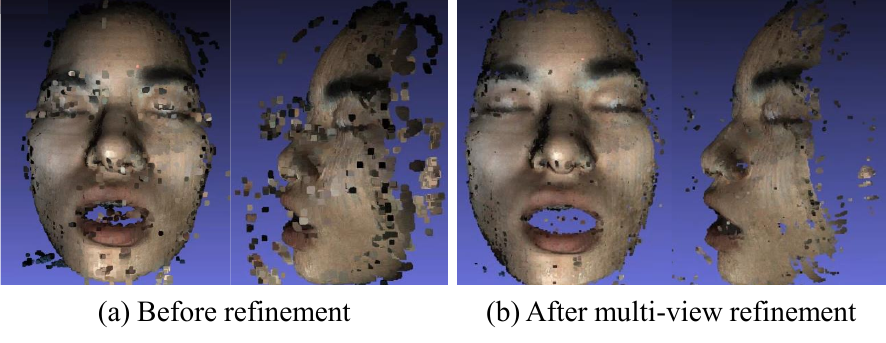}
\vspace{-3mm}
\caption{\textbf{Results of multi-view depth refinement.} We visualize point clouds (a) before refinement and (b) after multi-view refinement.}
\label{supplefig:pointrefinement}
\vspace{-2mm}
\end{figure}

Structured light can give us high-quality facial geometry that can be regarded as the ground-truth depth maps. 
There still exists outliers that mainly comes from small movement of face while capturing different coded patterns. 
To resolve this problem, we check the visibility of acquired points in perspective view of each camera and remove outliers of gathered point clouds with considering neighbor points. 
Finally, we check each points' multi-view photometric/depth consistency (each point should be visible from more than two sampled views) to determine whether the point is inlier or not. Results in~\Fref{supplefig:pointrefinement} shows that this refinement process can reduce outliers effectively without loosing inliers.

\subsection{Surface Normal from Photometric Stereo}
\label{subsec:supple_photometricstereo}
In this section, we will explain our progress to get good quality of normal map to be used for supervised signal in training step.
We use a photometric stereo as reported in~\cite{nehab2005efficiently} to estimate surface normal of the subjects' face. First, we capture multiple images with light from varying directions. We calibrate the lighting directions using a chrome ball and estimate them as follows:

\begin{equation}
\footnotesize
\begin{gathered}
\mathbf{L}=2(\mathbf{n}\cdot \mathbf{R})\mathbf{N} - \mathbf{R},
\text{where}~\mathbf{R}=(0,0,1)^\intercal, \mathbf{n}=\frac{1}{r}(n_x,n_y,n_z), \\
n_x=h_x-c_x, n_y=h_y-c_y, n_z=\sqrt{r^2-n_x^2-n_y^2},
\label{eq:light}
\end{gathered}
\end{equation}
where $c$ and $h$ are the center of a chrome ball, 
and the center of specular reflection, respectively. 
$r$ is the norm of the normal vector $(n_x,n_y,n_z)$. 
By finding the distance between the centers $c$ and $h$, 
the lighting direction $\mathbf{L}=(L_x, L_y, L_z)$ can be calculated using~\Eref{eq:light}. 

Then, we compute the surface normal of the subject by solving a large over-constrained linear system as below:
\begin{equation}
\footnotesize
\begin{bmatrix}
\mathbf{N}_1\\...\\\mathbf{N}_p
\end{bmatrix}=\begin{bmatrix}
\mathbf{I}_1\\...\\\mathbf{I}_p
\end{bmatrix}
[L_x, L_y, L_z]^\intercal,~~p\in \mathbf{P},
\end{equation}

Finally, we follow outlier-robust scheme~\cite{mukaigawa2007analysis,hernandez2008multiview} to reject non-Lambertian observations by regarding them as outliers.

\subsection{Normal-guided Depth Refinement}
\label{subsec:Normal-guided Depth Refinement}
We further continue to remove outliers and improve the quality of depth by constraining the surface gradient with given normal map from~\Sref{subsec:supple_photometricstereo}, which is orthogonal to the photometric normal while keeping its original position as possible. For this, we formulate an energy function consisting of two error terms: the position error $E_d$ and the normal error $E_n$:
\begin{equation}
\footnotesize
\begin{gathered}
E = \lambda E_d + (1-\lambda) E_n,\\
E_{d}=\left\|\mathbf{X}_{p}-\mathbf{X}_{p}^{m}\right\|, \\ E_{n}=\sum_{p}\left(\left[T_{x}\left(\mathbf{X}_{p}\right) \cdot \mathbf{N}_{p}\right]^{2}+\left[T_{y}\left(\mathbf{X}_{p}\right) \cdot \mathbf{N}_{p}\right]^{2}\right)
\label{eq:refine}
\end{gathered}
\end{equation}

\begin{figure}[!t]
\centering
\includegraphics[width=0.99\linewidth]{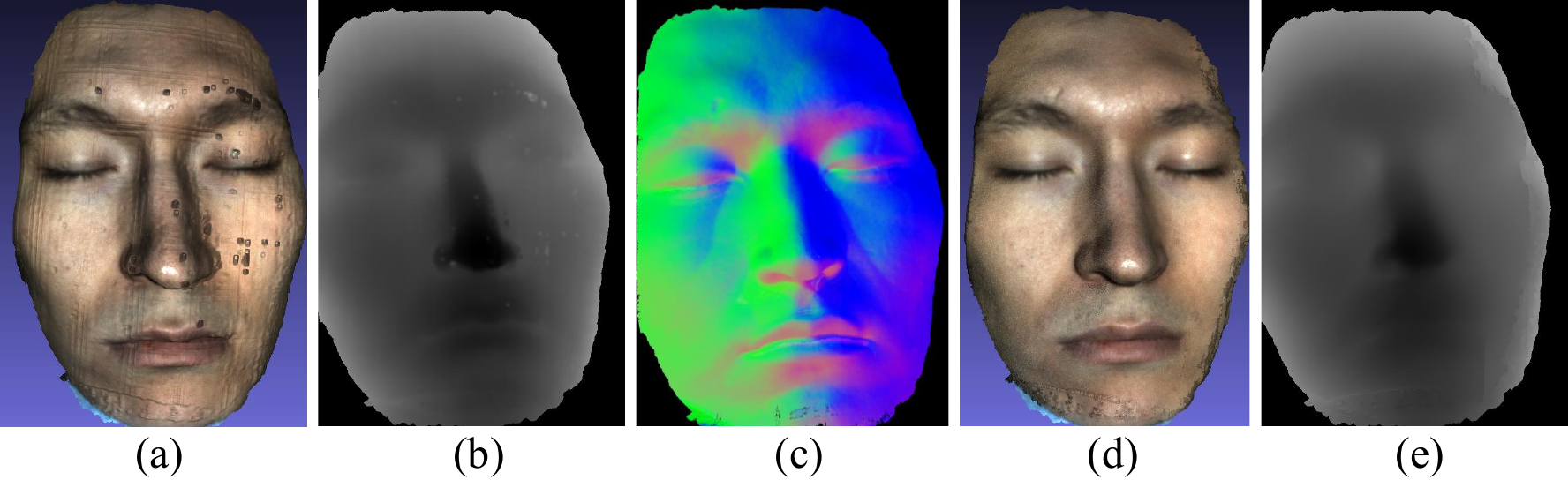}
\vspace{-2mm}
\caption{\textbf{Visualization of an example face of the normal-guided depth refinement.} (a)~3D mesh and (b)~depth map before refinement. (c)~An estimated normal map. (d)~Refined mesh and (e)~depth map after normal-guided depth refinement.}
\label{supplefig:surfacerefinement}
\vspace{-2mm}
\end{figure}

where $\mathbf{X}_p$, $T_{x,y}(\mathbf{X}_p)$ are a 3D point and its surface gradient in the $x,y$-direction, respectively. $\lambda$ is the balancing term between the position and the normal errors.  The entire minimization is also formulated as an over-constrained linear system to be solved by least squares as described in~\cite{nehab2005efficiently}. 

As shown in the~\Fref{supplefig:surfacerefinement}, the refinement process effectively removes outliers and improves the overall quality of facial geometry by alleviating the line-artifact and noisy 3D points.

\begin{figure}[!t]
\centering
\includegraphics[width=0.95\linewidth]{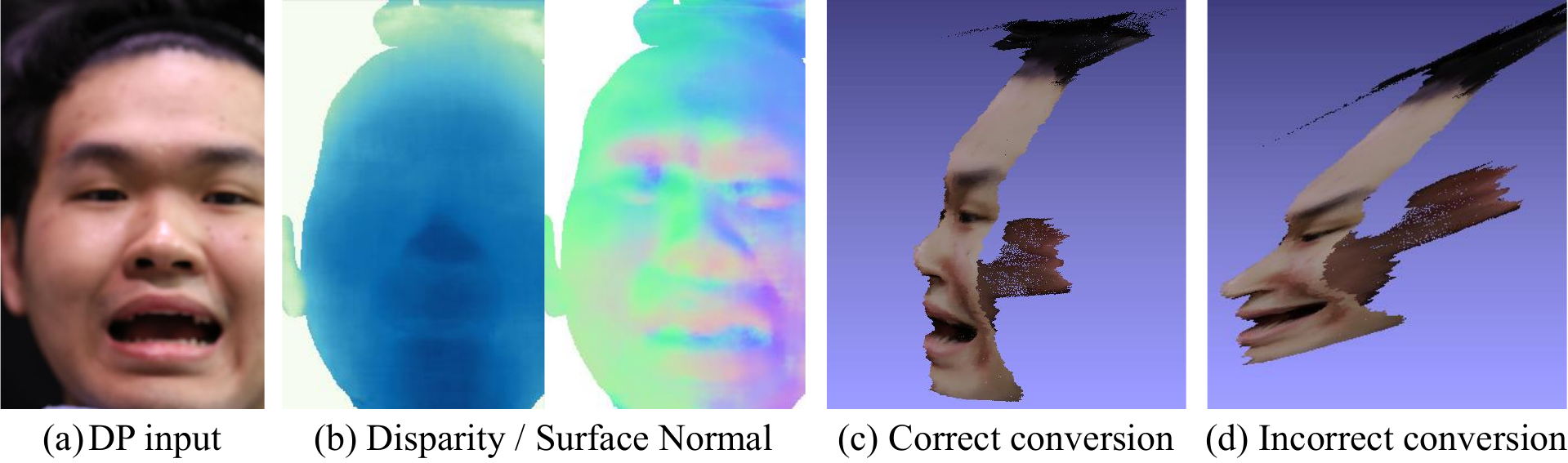}
\vspace{-2mm}
\caption{\textbf{Facial reconstruction from given facial DP image.} Given DP input of (a), our proposed StereoDPNet estimates disparity map and surface normal in (b). We show the reconstructed point cloud using correct/incorrect conversion of~\Sref{subsec:disp2depth_supple} in (c), (d) to show the importance of our calibration process.}
\label{supplefig:wrong_convert}
\end{figure}
\begin{figure}[!t]
\centering
\includegraphics[width=0.99\linewidth]{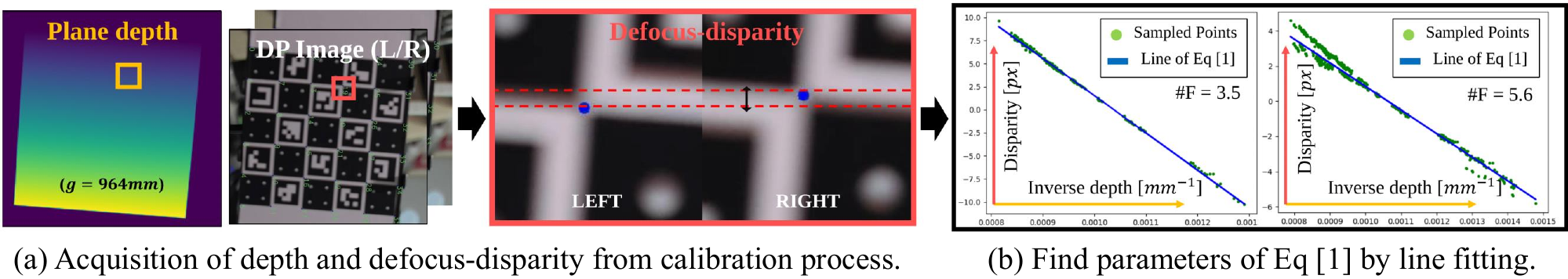}
\vspace{-2mm}
\caption{\textbf{Defocus-disparity to metric depth.} (a) We first acquire defocus-disparity and depth obtained by a plane homography. (b) Using the acquired depth and disparity, we find parameters of~\Eref{eq:disparity_to_depth}.}
\label{supplefig:calib}
\end{figure}

\subsection{Conversion from Disparity to Metric Depth}
\label{subsec:disp2depth_supple}
Given the estimated defocus-disparity from our proposed network, StereoDPNet, we provide exact conversion between the disparity and the metric depth in~\Sref{subsec:ground truth acquisition} in the manuscript. As shown in~\Fref{supplefig:wrong_convert}, finding the correct relationship is critical to facial 3D reconstruction since the wrong conversion can result twisted shape. Here, we explain our calibration process in details that covers conversion between a defocus-disparity and a metric-scale depth.

In~\Fref{supplefig:calib}~(\textcolor{red}{a}), 
we compute the corresponding pair of points of left/right DP images.
In particular, we adopt the saddle point refinement method~\cite{ha2017deltille} that is robust to defocus blur in DP images. 
By doing so, we obtain plane depth and defocus-disparity at each point.

In~\Fref{supplefig:calib}~(\textcolor{red}{b}), following~\Eref{eq:disparity_to_depth}, we use the linear relation between the obtained inverse depth and defocus-disparity  measurements.
To find calibration parameters, we first compute the bias~$B(L, f, g)$ and the slope~$A(L, f, g)$ through least-square optimization.
Then, we obtain the focus distance $g{=}{-}\frac{A(L, f, g)}{B(L, f, g)}$. 
Using focal length $f$ and F number that are pre-defined by the lens condition, we calculate the aperture $L$.
$\alpha$ is acquired from the slope~$A(L, f, g)$.
Finally, we get all the calibration parameters of~\Eref{eq:disparity_to_depth}.

\newpage

\section{Evaluation Metrics}
\label{sec:supple_metrics}
\noindent\textbf{Depth Metrics.}\quad
Previous studies~\cite{garg2019learning,punnappurath2020modeling} predict depth with affine ambiguity. Therefore, their methods only provide experimental results with affine invariant metrics.
Following~\cite{garg2019learning}, we measure the quality of affine transformed depth as:

\begin{enumerate}
{\setdefaultleftmargin{3.2mm}{}{}{}{}{}
    \item[$\bullet$] Affine invariant metrics
    \begin{itemize}
    \footnotesize
        \item[-] AIWE(p) : $\underset{a, b}{\operatorname{min}}\left(\frac{\sum_{u=1}^W\sum_{v=1}^H\left|d_{u,v}-(a \hat{d}_{u,v}+b)\right|^{p}}{|H \cdot W|}\right)^{1 / p}$
        \item[-] WMAE = AIWE(1)
        \item[-] WRMSE = AIWE(2)
    \end{itemize}
}
\end{enumerate}

Furthermore, based on our calibration parameters (\Sref{subsec:disp2depth_supple}), we are able to measure the accuracy of absolute-scale depth\footnote{\url{http://www.cvlibs.net/datasets/kitti/}} as:
\begin{enumerate}
{\setdefaultleftmargin{3.2mm}{}{}{}{}{}
    \item[$\bullet$] Absolute metrics
    \begin{itemize}
    \footnotesize
        \item[-] RMSE : $\sqrt{\frac{1}{|H \cdot W|} \sum_{u=1}^W\sum_{v=1}^H\left|Z_{u,v}-\hat{Z}_{u,v}\right|^{2}}$
        \item[-] AbsRel : $\frac{1}{|H \cdot W|} \sum_{u=1}^W\sum_{v=1}^H\left|\left(Z_{u,v}-\hat{Z}_{u,v}\right) / Z_{u,v}\right|$
        \item[-] MAE : $\frac{1}{|H \cdot W|} \sum_{u=1}^W\sum_{v=1}^H\left|Z_{u,v}-\hat{Z}_{u,v}\right|$
        \item[-] $\delta^{i}$ : $\frac{1}{|H \cdot W|} \sum_{u=1}^W\sum_{v=1}^H \Big(\max \left(\frac{Z_{u,v}}{\hat{Z}_{u,v}}, \frac{\hat{Z}_{u,v}}{Z_{u,v}}\right)<\tau^{i}$\Big)
    \end{itemize}
}
\end{enumerate}
where $\hat{Z}$ denotes estimated depth and $Z$ denotes ground-truth depth.
Here, we used $\tau$ as $1.01$ where $i \in$ $\{1, 2, 3\}$. 

\noindent\textbf{Normal Metrics.}\quad
Following~\cite{kusupati2020normal}, we use Mean Angular Error (MAE) and Root Mean Square Angular Error (RMSAE) as:

\begin{enumerate}
{\setdefaultleftmargin{3.2mm}{}{}{}{}{}
    \item[$\bullet$] Normal metrics
    \begin{itemize}
    \footnotesize
        \item[-] MAE : $\frac{1}{|H \cdot W|}\sum_{u=1}^W\sum_{v=1}^H \arccos \left(\mathbf{n}_{u,v} \cdot \hat{\mathbf{n}}_{u,v}\right)$
        \item[-] RMSAE : $\sqrt{\frac{1}{|H \cdot W|}\sum_{u=1}^W\sum_{v=1}^H {\arccos \left(\mathbf{n}_{u,v} \cdot \hat{\mathbf{n}}_{u,v}\right)}^{2}}$
    \end{itemize}
}
\end{enumerate}

\newpage 

\section{Details of Adaptive Normal Module}
\label{sec:supple_algorithm}
In this paper, we propose two sub-modules for predicting depth and surface normal from a single pair of facial DP images.
Here, we describe details of surface sampling layer of our proposed ANM in~\Aref{alg:anm_1}, and \Aref{alg:anm_2}.

\begin{algorithm}[!h]
\footnotesize
\caption{Surface Sampling in ANM}
\label{alg:anm_1}
\begin{algorithmic}[1]

\Require Aggregated Cost Volume, $C_{\text{A}} \in \mathbb{R}^{C{\times}M{\times}H{\times}W}$
\begin{itemize}
  \item[]~~~ Inferred disparity, $\hat{d} \in \mathbb{R}^{H{\times}W}$
  \item[]~~~ Number of sampled neighbors, $P{=}4$
  \item[]~~~ Number of hypothesis planes in~$C_{\text{A}}$, $M{=}8$
\end{itemize}

%

\Procedure{SurfaceSample}{$C_{\text{A}}$,~$\hat{d}$,~$P$,~$M$}
    \State $\tilde{d}_{\text{P}} \leftarrow$Convert-Disparity-to-VolumeIndex($\hat{d}, M$) \Comment{Alg.~2}
    \State $C_{S}\leftarrow$Sample-$P$-Closest-Neighbors($C_{A}, \tilde{d}_{\text{P}}, P, M$)
    \State \Return Sampled volume~$C_{S} \in \mathbb{R}^{C\times P\times H\times W}$
\EndProcedure
\end{algorithmic}
\end{algorithm}
\vspace{-4mm}


\begin{algorithm}[!h]
\footnotesize
\caption{Ray Sampling in Volume}
\label{alg:anm_2}
\begin{algorithmic}[1]
\Require Inferred disparity $\hat{d} \in  \mathbb{R}^{H{\times}W}$
\begin{itemize}
  \item[]~~~ Number of hypothesis planes in~$C_{\text{A}}$, $M{=}8$
  \item[]~~~ $d_\text{min} = d^0$, $d_\text{max} = d^m$ (Eq.~2 of the manuscript)
  \item[]~~~ VolumeIndex $\tilde{d}_{\text{P}} \in \mathbb{R}^{3\times H\times W}$ 
\end{itemize}

\Procedure{Convert-Disparity-to-VolumeIndex}{$\hat{d}$,~$M$}
    \For{$(u, v)$ in $\hat{d}$}
        
        \State $\tilde{d}_{\text{P}(u,v)} \leftarrow \big(u, v, \frac{\hat{d}_{(u,v)} - d_\text{min}}{d_\text{max} - d_\text{min}} \cdot M \big)$
    \EndFor
    \State \Return VolumeIndex $\tilde{d}_{\text{P}} \in \mathbb{R}^{3\times H\times W}$ 
\EndProcedure


\end{algorithmic}
\end{algorithm}
\vspace{-4mm}

\newpage

\section{Supplementary Results}
\label{sec:supple_results}

\subsection{Comparison of DPNet with the Original}
\label{subsec:compare_dpnet_supple}
Since there is no public code for DPNet~\cite{garg2019learning}, we re-implement DPNet~\cite{garg2019learning} by ourselves following their description. 
To verify our implementation, we train our DPNet and measure the performance on their dataset~\cite{garg2019learning}. 
Although there is a little performance drop with our implementation compared to the reported performance in the original paper~\cite{garg2019learning}, we show that our implemented model has similar performance with the original implementation in~\Tref{tab:supple_dpnet}.

\vspace{-4mm}
\begin{table}[H]
\centering
\scriptsize
\setlength{\tabcolsep}{10pt}
\begin{tabular}{cccc}
\\ \Xhline{4\arrayrulewidth}
\multirow{2}{*}{Method} & \multicolumn{3}{c}{Affine error metric $\downarrow$} \\
                        & AIWE(1)     & AIWE(2)     & 1 - $\rho$     \\ \Xhline{2\arrayrulewidth}
DPNet (reported in~\cite{garg2019learning})    & 0.0581      & 0.0735      & 0.827       \\ \Xhline{1\arrayrulewidth}
DPNet (reimplemented)   & 0.073       & 0.09        & 0.883      \\ \Xhline{4\arrayrulewidth}
\end{tabular}
\vspace{1mm}
\caption{\textbf{Comparison of DPNet reported in~\cite{garg2019learning} and re-implemented by ours.} We measure the performance of our re-implemented DPNet on their dataset~\cite{garg2019learning} and compare with the reported performance in~\cite{garg2019learning}. Note that we don't use their affine invariant loss to purely verify the performance of the model.}
\label{tab:supple_dpnet}
\vspace{-4mm}
\end{table}

\subsection{Application : Face Relighting}
Although our main goal is to estimate the depth and normal from DP images, we introduce naive methods for one of the applications, face relighting, that can be the baseline for future works.

\begin{figure}[!t]
\centering
\includegraphics[width=0.95\linewidth]{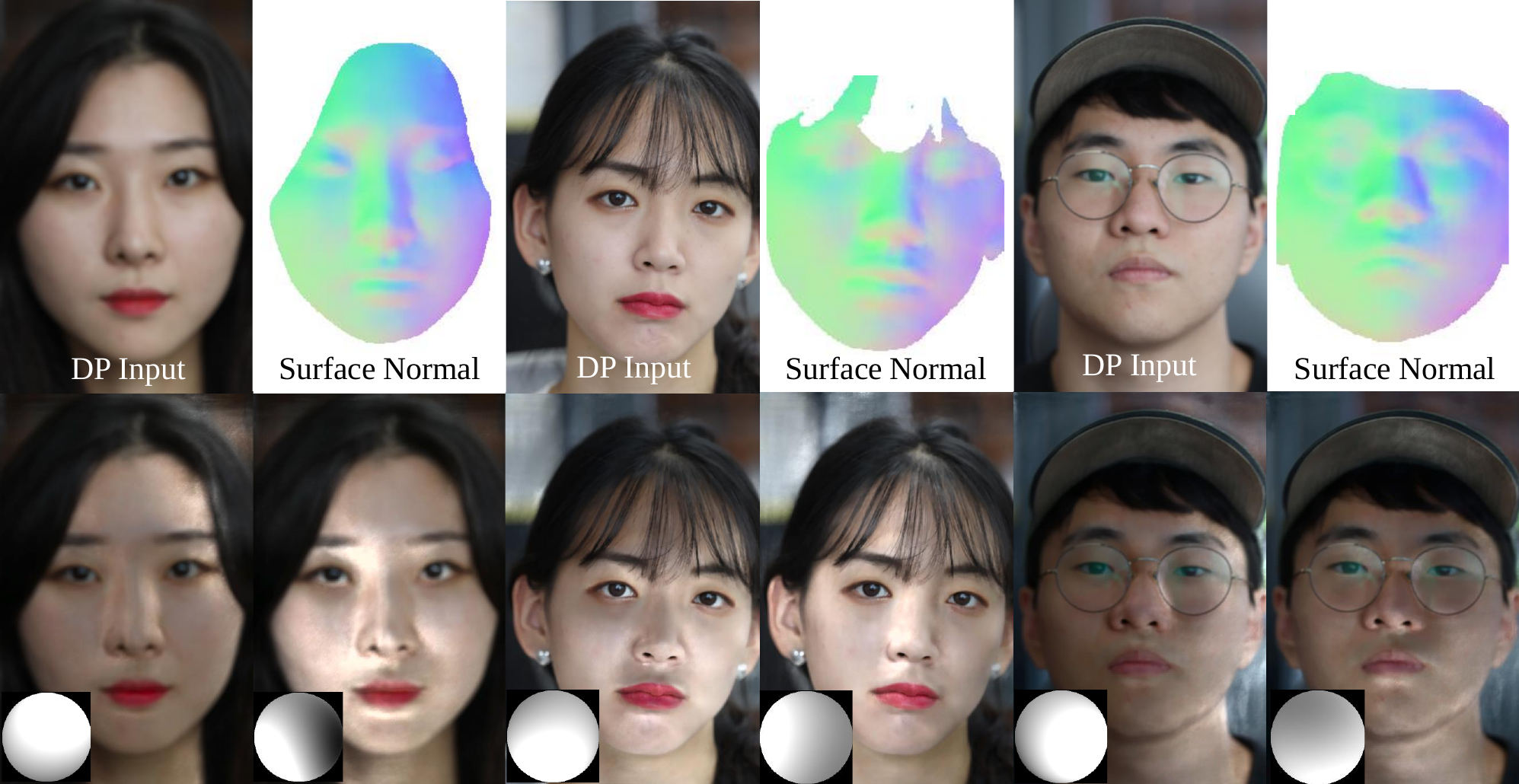}
\vspace{-3mm}
\caption{\textbf{Face relighting from our estimated surface normal.} We display the original DP image, estimated normal map from StereoDPNet, and two different relit images with sampled lighting directions.}
\label{supplefig:relighting}
\vspace{-3mm}
\end{figure}

\noindent\textbf{Face Relighting.}\quad
Given the reference images and the surface normals from StereoDPNet, we generate relighted images using a Ratio Image-based method \cite{zhou2019deep}. The target spherical harmonic lightings are randomly sampled and the lighting directions of reference images are inferred with SfSNet~\cite{sengupta2018sfsnet}. The results are shown in~\Fref{supplefig:relighting}.

\subsection{Real-World Results}
\label{subsec:realworld_supple}
\begin{figure}[!t]
\centering
\includegraphics[width=0.95\linewidth]{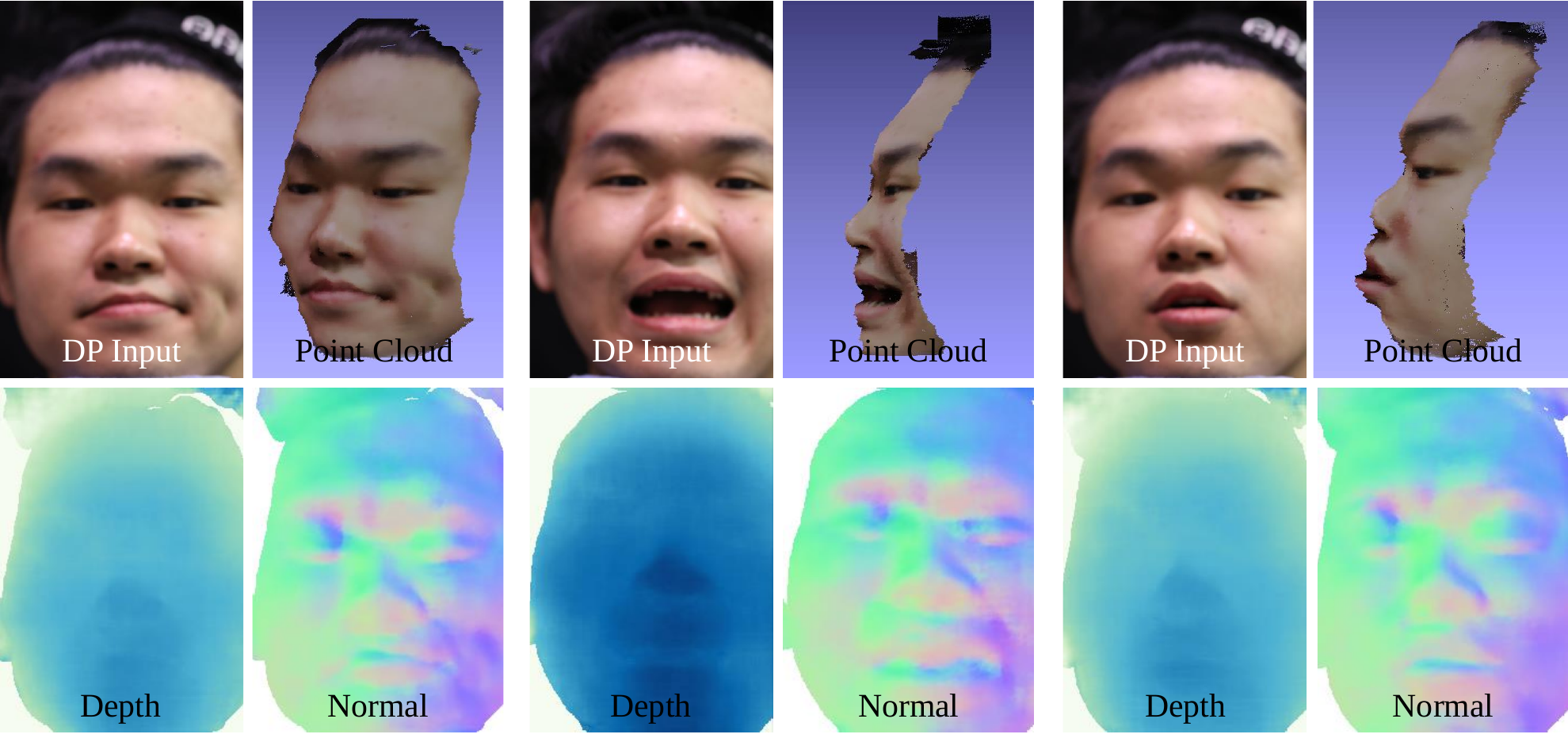}
\vspace{-3mm}
\caption{\textbf{Depth and Normal with facial expressions.} We demonstrate that our method can cover face with various facial expressions thanks to our carefully designed facial dataset.}
\label{supplefig:face_expression}
\vspace{-3mm}
\end{figure}

\begin{figure}[!t]
\centering
\includegraphics[width=0.99\linewidth]{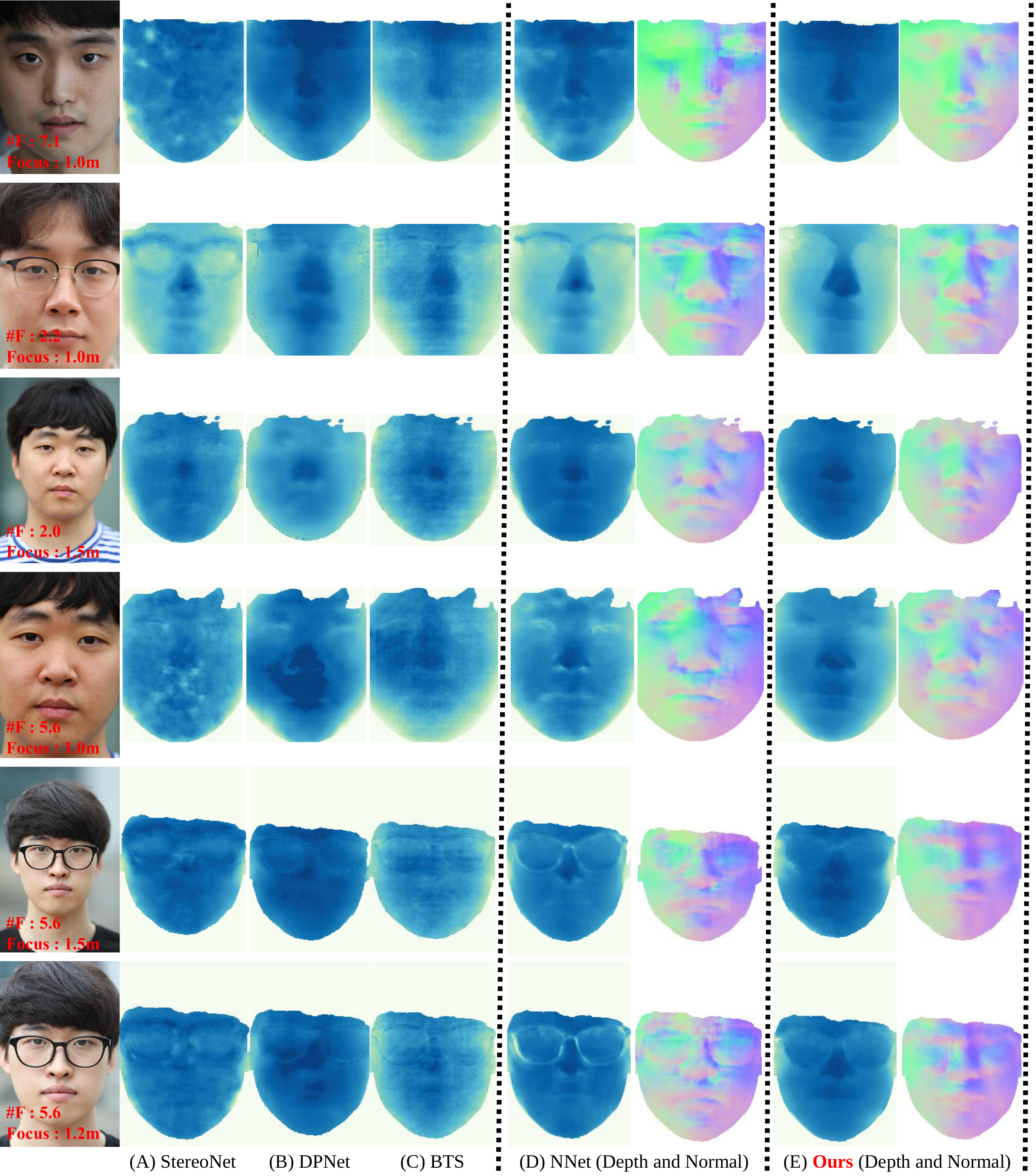}
\vspace{-2mm}
\caption{\textbf{Real-world results.} More real-world results with captured camera settings similar to~\Fref{fig:real_world_result}.}
\label{supplefig:realworld1}
\end{figure}

\begin{figure}[!t]
\centering
\includegraphics[width=0.99\linewidth]{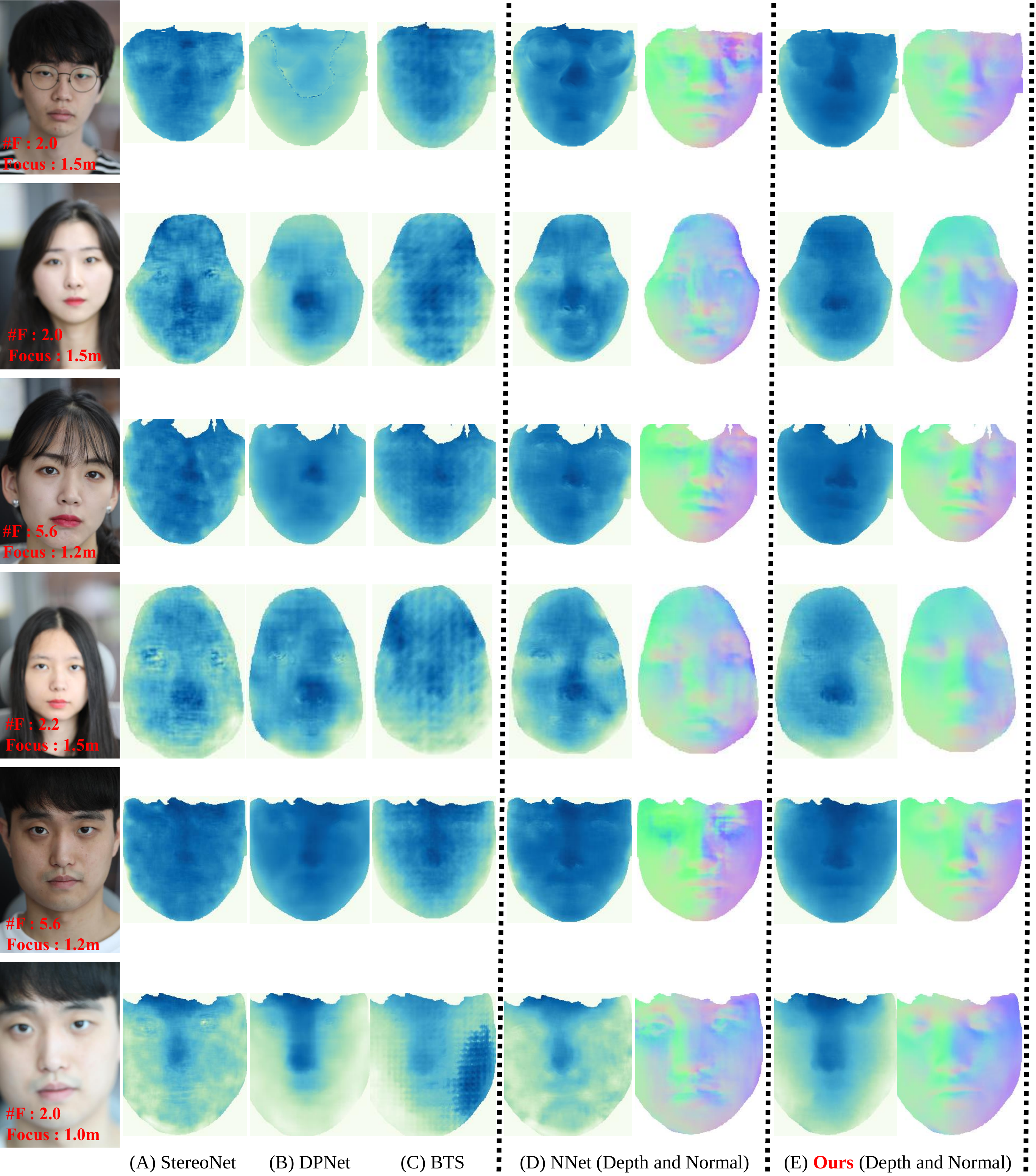}
\vspace{-2mm}
\caption{\textbf{Real-world results.} More real-world results with captured camera settings similar to~\Fref{fig:real_world_result}.}
\label{supplefig:realworld2}
\end{figure}

We show additional real-world results with unmet environment to demonstrate our method and dataset's generality in~\Fref{supplefig:realworld1} and in~\Fref{supplefig:realworld2} similar to~\Fref{fig:real_world_result} of manuscript. We also note that our method is able to apply with various camera parameters (focus distance from 1.0m to 1.5m and F-number from 2.0 to 7.1). We also demonstrate that our method can deal with various facial expressions in~\Fref{supplefig:face_expression}.



\clearpage
%
%
\bibliographystyle{splncs04}
\bibliography{egbib}
\end{document}